\documentclass[lettersize,journal]{IEEEtran}
\usepackage{amsmath,amsfonts}
\usepackage[ruled]{algorithm2e}
\usepackage{array}
\usepackage{multirow}
\usepackage{amssymb}
\usepackage[caption=false,font=normalsize,labelfont=sf,textfont=sf]{subfig}
\usepackage{lipsum} % For dummy text
\usepackage{adjustbox} % For centering the table
\usepackage{multirow} % For multirow cells

\usepackage{textcomp}
\usepackage{stfloats}
\usepackage{adjustbox} % For centering the table
\usepackage{hyperref}
\usepackage{booktabs}
\usepackage{verbatim}
\usepackage{graphicx}
\usepackage{hyperref}
\usepackage{xcolor}
\usepackage{cite}
\begin{document}

\title{CLIP-based Camera-Agnostic Feature Learning for Intra-camera Person Re-Identification}

% CLIP-based Multi-label Adversarial Learning for Intra-camera Supervised Person Re-Identification

% CLIP-based Camera-Aware Adversarial Learning for Intra-camera Person Re-Identification

% CLIP-based Adversarial Camera Alignment Network for Intra-camera Supervised Person Re-Identification

\author{
Xuan~Tan,
Xun~Gong,~\IEEEmembership{Member,~IEEE},
and~Yang~Xiang
 
\thanks{. \textit{}

Xuan Tan and Yang Xiang are with the Tangshan Research Institute, Southwest Jiaotong University, Tangshan 063000, China 
(e-mail: trangle@my.swjtu.edu.cn; xiangyang@my.swjtu.edu.cn).

Xun Gong is with the School of Computing and Artificial Intelligence, Southwest Jiaotong University, Chengdu, Sichuan 610031, China, Engineering Research Center of Sustainable Urban  Intelligent Transportation, Ministry of Education, China, and also with Manufacturing Industry Chains Collaboration and Information Support Technology Key Laboratory of Sichuan Province, Chengdu, Sichuan 610031, China(e-mail: xgong@swjtu.edu.cn).}}

% \IEEEpubid{\begin{minipage}{\textwidth}\ \centering
% 		Copyright \copyright 20xx IEEE. Personal use of this material is permitted. \\
% 		However, permission to use this material for any other purposes must be obtained 
% 		from the IEEE by sending an email to pubs-permissions@ieee.org.
% \end{minipage}}

\maketitle

\begin{abstract}
Contrastive Language-Image Pre-Training (CLIP) model excels in traditional person re-identification (ReID) tasks due to its inherent advantage in generating textual descriptions for pedestrian images. 
However, applying CLIP directly to intra-camera supervised person re-identification (ICS ReID) presents challenges. ICS ReID requires independent identity labeling within each camera, without associations across cameras. This limits the effectiveness of text-based enhancements. 
To address this, we propose a novel framework called CLIP-based Camera-Agnostic Feature Learning (CCAFL) for ICS ReID.
Accordingly, two custom modules are designed to guide the model to actively learn camera-agnostic pedestrian features: Intra-Camera Discriminative Learning (ICDL) and Inter-Camera Adversarial Learning (ICAL). Specifically, we first establish learnable textual prompts for intra-camera pedestrian images to obtain crucial semantic supervision signals for subsequent intra- and inter-camera learning. Then, we design ICDL to increase inter-class variation by considering the hard positive and hard negative samples within each camera, thereby learning intra-camera finer-grained pedestrian features.
Additionally, we propose ICAL to reduce inter-camera pedestrian feature discrepancies by penalizing the model's ability to predict the camera from which a pedestrian image originates, thus enhancing the model's capability to recognize pedestrians from different viewpoints.
Extensive experiments on popular ReID datasets demonstrate the effectiveness of our approach. Especially, on the challenging MSMT17 dataset, we arrive at 58.9\% in terms of mAP accuracy, surpassing state-of-the-art methods by 7.6\%.
Code will be available at: https://github.com/Trangle12/CCAFL.
\end{abstract}


\section{Introduction}
\IEEEPARstart{P}{erson} re-identification (Re-ID) involves identifying the same individual across different camera views. It has attracted significant attention because of its applications in person tracking, security systems, and traffic monitoring. Current research primarily focuses on two directions: fully supervised \cite{chen2019abd},\cite{sun2019learning},\cite{luo2019bag},\cite{he2021transreid},\cite{li2021diverse} and unsupervised \cite{ge2020self},\cite{ge2020mutual},\cite{wang2022offline},\cite{zhu2022pass}. With the advent of deep learning technologies, fully supervised person ReID has seen significant performance improvements. However, the considerable annotation cost associated with the increasing number of cameras and IDs in real-world scenarios poses a significant challenge for the practical deployment of ReID systems. Conversely, unsupervised person ReID does not require any label information but tends to underperform in complex scenarios involving multiple IDs.
In recent years, to combine the advantages and mitigate the drawbacks of supervised and unsupervised methods, the Intra-camera supervision (ICS) approach has been proposed.

\begin{figure}[t]
  \centering
   \includegraphics[width=0.45\textwidth]{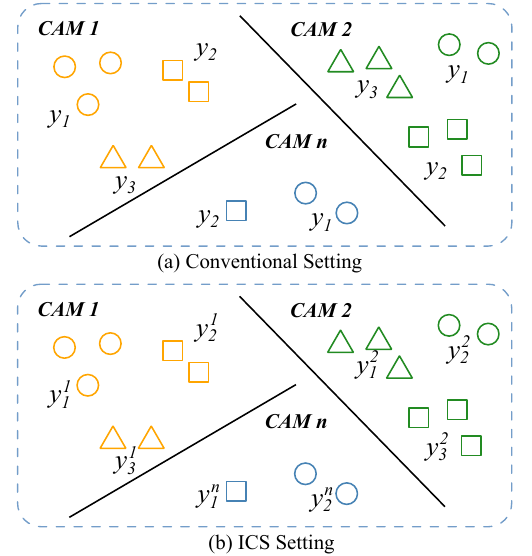}
  \caption{Illustration of label settings under different person Re-ID data configurations. The light-blue areas represent the intra-camera and cross-camera feature spaces, with different shapes corresponding to different identities. (a) Conventional fully supervised training data requires unified identity annotation across all cameras. (b) Intra-camera supervised (ICS) training data only requires independent identity annotation within each camera view, utilizing separate class spaces. In ICS ReID data, superscripts of identity labels indicate camera view labels.}
  \label{fig:ics}
\end{figure}

This approach assumes individual labeling of IDs within each camera without establishing cross-camera identity links. 
As a result, ICS supervision significantly reduces annotation costs compared to full supervision while still maintaining identification accuracy.
Therefore, it is considered a more practical setup for ReID scenarios.

However, the lack of cross-camera annotation information poses a significant challenge for effectively learning pedestrian features in ICS ReID. 
Specifically, the number of annotated pedestrian training samples within each camera is significantly lower than in fully supervised cross-camera person ReID tasks. 
Additionally, due to factors such as varying viewpoints, occlusion, and background noise, the absence of inter-camera labels makes it difficult for models to learn variations in pedestrian appearance across different views, as illustrated in Fig. \ref{fig:ics}. 
Therefore, effectively utilizing intra-camera supervised information to learn associations between cross-camera IDs is crucial for addressing ICS ReID tasks.

\begin{figure}[t]
  \centering
   \includegraphics[width=0.450\textwidth]{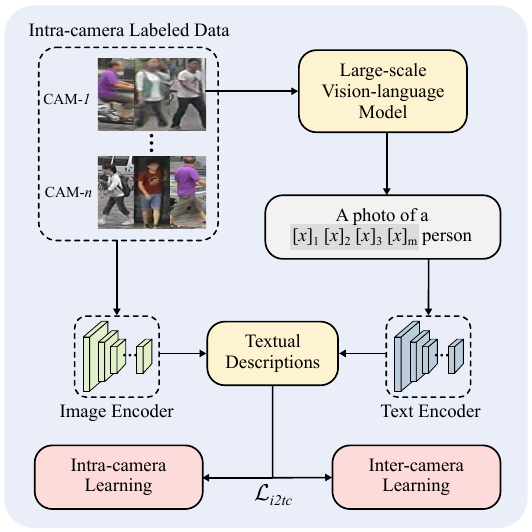}
  \caption{The diagram illustrates our proposed approach, which leverages CLIP and prompt learning to generate textual descriptions for person images within each camera.  Based on this, we combine the textual information with intra-camera and inter-camera learning, enabling the model to focus better on discriminative features.}
  \label{fig:flow}
\end{figure}

In ICS ReID, a common approach is to divide model learning into two stages: intra-camera learning and inter-camera learning. For instance, the multi-label learning strategy MATE  \cite{zhu2021intra} constructs a Softmax parameter classifier for each camera to classify pedestrians while associating cross-camera identity labels. However, the variation in the number of pedestrian samples within each camera may lead to suboptimal performance. To address this issue, Precise-ICS \cite{wang2021towards} constructs a non-parametric classifier \cite{xiao2017joint} for each camera and continues to train the model by assigning pseudo-labels to highly similar cross-camera images.
Despite these advancements, such methods fail to fully utilize intra-camera sample annotation information, and pseudo-labels obtained through simple clustering of different camera angles are often inaccurate and lack flexibility. Notably, the recent large-scale vision-language model CLIP \cite{radford2021learning} has demonstrated inherent advantages in generating textual descriptions for images. We can leverage CLIP to describe unseen pedestrian features, thus generating general descriptions of new pedestrian images without additional annotation data. This allows us to use the semantic information generated by CLIP to assist in correcting label associations during the inter-camera learning phase while distinguishing features of different pedestrians.

In this paper, we propose a CLIP-based Camera-Agnostic Feature Learning (CCAFL) method to address the ICS ReID problem through a three-stage learning process. As shown in Fig. \ref{fig:flow}, our method's components work together to tackle the challenges of ICS ReID. First, in the initial stage, we employ prompt learning to generate textual descriptions for individual images within each camera. These textual descriptions provide additional supervisory information for subsequent intra-camera and inter-camera learning stages.

During the intra-camera learning stage, to further enhance the discriminative power of pedestrian features within each camera, we construct an independent hybrid feature memory bank for each camera using annotated intra-camera IDs. This memory bank stores the average features and all instance features for each ID. Subsequently, we apply Intra-Camera Discriminative Learning (ICDL), which considers hard-to-classify positive and negative samples within the same camera while leveraging textual features obtained in the first stage to further reduce intra-camera intra-class variance and inter-class similarity.

In the inter-camera learning stage, to improve the accuracy of cross-camera pedestrian ID associations and enhance the model's ability to recognize pedestrians from different viewpoints, we first use a cross-camera association algorithm to link cross-camera pedestrian IDs. Using these associated IDs, we construct a cross-camera feature memory bank that stores their central features for contrastive learning. Then, utilizing the intra-camera supervisory semantic information obtained previously, we assign a textual description to each cluster based on the generated pseudo-labels. These textual descriptions effectively summarize individual images and serve as additional supervisory information for inter-camera learning. Finally, to better reduce the data distribution differences between different camera views, we propose an Inter-Camera Adversarial Learning (ICAL) method. Specifically, we add a multi-camera classifier after the backbone network of the re-identification model and define ICAL as a multi-positive class classification loss. During training, by minimizing ICAL, we force the backbone network of the re-identification model to learn camera-invariant features by penalizing the model's ability to predict the corresponding camera for the same identity. Through backpropagation, the learned feature maps can highlight more camera-invariant features.

Our main contributions can be summarized as follows:
\begin{itemize}
\item
We propose a simple yet effective three-stage training strategy, called CCAFL, that integrates CLIP-generated textual information into the novel semi-supervised ICS ReID task for subsequent learning.
% \item
% We propose an Intra-Camera Discriminative Loss (ICDL) that fully utilizes annotated samples within each camera to effectively improve the model's ability to distinguish pedestrian identities within the same camera.
% \item
% We propose an Inter-Camera Adversarial Loss (IAL) to reduce the Inter-camera feature distribution differences. By utilizing this loss, we compel the model to learn camera-agnostic features, thereby enhancing the accuracy of cross-camera pedestrian identity recognition.
\item 
Two critical modules: Intra-Camera Discriminative Learning (ICDL) and Inter-Camera Adversarial Learning (ICAL) are introduced to compel the model to learn camera-agnostic features. ICDL aims to extract intra-camera fine-grained pedestrian features, while ICAL reduces the inter-camera discrepancies in pedestrian feature distribution. These modules collectively enhance the accuracy of cross-camera pedestrian identity recognition.
\item
Extensive experiments conducted on three popular person re-identification benchmarks, Market-1501 \cite{zheng2015scalable} and MSMT17 \cite{wei2018person}, demonstrate that our method significantly outperforms the current state-of-the-art ICS methods. Our performance even exceeds that of fully supervised methods.
\end{itemize}

\section{Related Work}
\subsection{Intra-camera Supervised ReID}
With the increasing number of cameras and persons in real-world scenarios, the task of annotating a large-scale ID dataset becomes prohibitively costly. To address this issue, a setup known as Intra-camera supervision (ICS) ReID has been proposed, where annotations are performed independently across various cameras, with labels only available for persons within the same camera view.
Previous methods approached the ICS ReID problem from two angles: intra-camera supervised learning and inter-camera ID association learning. In intra-camera learning, PCSL \cite{qi2020progressive} and ACAN \cite{qi2021adversarial} employ a direct triplet loss \cite{hermans2017defense} to train models, while MATE \cite{zhu2021intra} constructs a multi-branch classifier for each camera. However, when the distribution of intra-camera ID samples is unbalanced and scant, it can result in biased learning. In contrast, Precise-ICS \cite{wang2021towards} uses a non-parametric classifier and undertakes joint learning, but insufficient intra-camera learning can severely impair the model when persons with high intra-camera similarity are present. For inter-camera learning, Precise-ICS supervises learning through pseudo-labeling based on the similarity of person features across cameras. CMT \cite{10534060} combines contrastive learning with the Mean Teacher \cite{tarvainen2017mean} paradigm to construct a semi-supervised learning framework. However, the methods above overlook the labeled instance features within the same camera, leading to insufficient intra-camera learning and consequently failing to effectively distinguish pedestrian features within the same camera. Additionally, these methods do not adequately consider the disparities in data distribution across different cameras, failing to fully capture the invariant features of pedestrians across cameras.

Differently, we design Intra-Camera Discriminative Loss (ICDL) and Inter-Camera Adversarial Loss (ICAL) methods to effectively enhance the model's ability to distinguish pedestrian ID features within and across cameras. Additionally, we incorporate high-level semantic features generated by CLIP for each person within a camera to further boost the model's performance.

\subsection{Unsupervised Person ReID}
In recent years, unsupervised person ReID \cite{ge2020self},\cite{ge2020mutual},\cite{wang2022offline},\cite{wang2021camera},\cite{cheng2022hybrid},\cite{dai2022cluster},\cite{chen2021ice} tasks have attracted wide attention. These tasks are primarily categorized based on whether additional related data are employed, encapsulating unsupervised domain-adaptive (UDA) ReID and purely unsupervised learning (USL) ReID. The latter, pure unsupervised ReID, presents greater challenges due to its independence from any external data. However, with the successful application of contrastive learning in the unsupervised domain, the performance of USL ReID has significantly increased - notable methods include SPCL \cite{ge2020self}'s self-paced contrastive learning procedure that builds a mixed memory bank, fully exploiting all available data. CAP \cite{wang2021camera} technique divides clusters into multiple camera-perception proxies based on the camera ID to alleviate discrepancies in ID features generated by camera perspective alterations. ClusterContrast \cite{dai2022cluster} directly establishes a simple yet effective cluster-level memory bank, achieving decoupling between feature update rates and the number of images. RTMem \cite{10102757} employs a real-time memory update strategy, updating cluster centroids by randomly sampling current mini-batch instance features without the need for momentum. In contrast, LP \cite{10137428} considers two types of additional features from different local views and leverages the knowledge of an offline teacher model to optimize the model. 
In this study, our work is grounded in the framework of intra- and inter-camera contrastive learning, which is a widely used and effective representation learning method for unsupervised person ReID.

\subsection{Vision-language Models}
Large-scale pre-trained vision-language models, integrating copious amounts of textual and visual data, have proven their efficacy across various domains. For example, Contrastive Language–Image Pretraining (CLIP) \cite{radford2021learning} model, which employs the InfoNCE loss \cite{he2020momentum} function to jointly train text and image encoders, resulting in significant performance improvements in numerous downstream tasks. Additionally, to further tap into CLIP's potential, CoOp \cite{zhou2022learning} introduces prompt learning, aiming to uncover the implicit textual cues within images, effectively migrating CLIP to a broader range of downstream tasks. Within the realm of ReID, CLIP has been extensively applied. For instance, CLIP-ReID \cite{li2023clip}, by aligning image and textual information within a singular embedding space, reinforces the connection between image features and related textual descriptions. 
CCLNet \cite{chen2023unveiling} establishes learnable cluster-aware prompts for person images and generates text descriptions to assist subsequent unsupervised visible-infrared person re-identification training.

However, the immense potential of CLIP in facilitating semi-supervised person ReID learning has yet to be explored. In this paper, we fully integrate CLIP with the ICS ReID task to construct a CCAFL framework, offering new insights for semi-supervised ReID.

\begin{figure*}[h]
  \centering
    \includegraphics[width=1.0\linewidth]{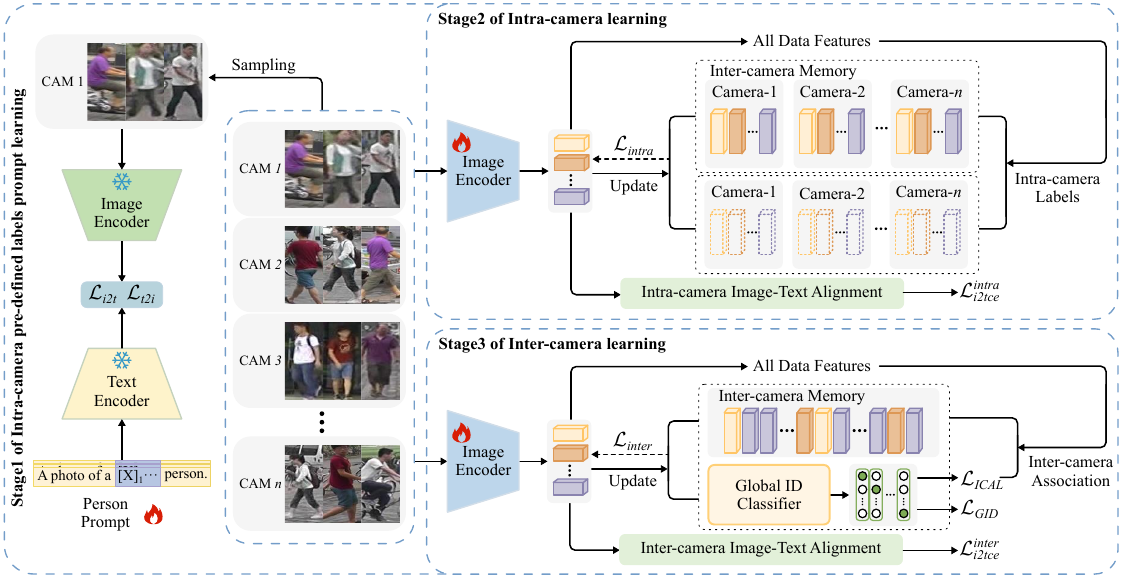}
  \caption{The framework of our CCAFL. 
  Left: Through prompt learning paradigms, we generate text descriptions corresponding to the labels of each person's image within a camera. This provides semantic supervision information for subsequent intra-camera and inter-camera learning. 
  Upper: In the intra-camera learning phase, we construct a hybrid memory for each camera, storing both the central features and instance features of pedestrians. By employing an intra-camera discriminative loss, we enhance the discriminability of pedestrian features within the same camera.
  Lower: In the inter-camera learning phase, we obtain cross-camera association IDs through a cross-camera association step. We then build a memory that stores prototype features of associated pedestrians, aiding the model in learning pedestrian features across different cameras. Additionally, we introduce a global ID classifier and incorporate inter-camera adversarial learning to mitigate the impact of camera discrepancies.
  }
  \label{fig:overview}
\end{figure*}

\subsection{Adversarial Learning}
The application of adversarial learning in person re-identification can be traced back to the use of Generative Adversarial Networks (GANs) \cite{goodfellow2014generative} to generate realistic person images. For example, \cite{jiang2021exploring} proposed a GAN-based method that performs selective sampling of generated data to bridge the gap between domains and enrich the feature space. In recent years, the application of adversarial learning has extended beyond image generation and has been widely applied to various aspects of person re-identification. 
For instance, in unsupervised domain adaptation for person re-identification, CAWCL \cite{mekhazni2023camera} employs a Gradient Reversal Layer (GRL) \cite{ganin2015unsupervised} to align the distribution of each camera. However, using traditional domain adversarial learning to eliminate camera styles can negatively impact the model's ability to recognize pedestrians.
In clothing change person re-identification, CAL \cite{gu2022clothes} proposed a clothing-based adversarial loss to decouple clothing-independent features. In contrast to these methods, we propose an inter-camera adversarial loss that penalizes the model's ability to predict the same identity under different cameras, thereby enabling the model to extract camera-agnostic features.

\section{METHODOLOGY}

\subsection{Overview}
Based on the ICS ReID problem, the training dataset only contains intra-camera IDs and lacks inter-camera IDs. Therefore, in this case, a dataset consisting of $C$ cameras can be represented as $D = \{D_1, D_2, \ldots, D_C\}$. Specifically, the images of persons from the $c$-th camera can be represented as ${D_c} = \{ ({x_i},{y_j},c)\}$, where ${x_i}$ indicates the $i$-th person image under this camera, ${y_j}(0 \le j < {N_c})$ represents the corresponding label, and ${N_c}$ denotes the number of person IDs under this camera. For instance, in the Market-1501 \cite{zheng2015scalable} dataset, which contains six cameras, there are 751 pedestrian IDs under supervised conditions, meaning these IDs are associated across different cameras. However, in the ICS setting, the IDs for each camera are independently annotated, with the number of pedestrian IDs for each camera being $D = \{ {\rm{652, 541, 694, 241, 576, 558}} \}$, resulting in a total of 3,262 global IDs. Although multiple cameras may capture the same pedestrian, their global IDs remain distinct.
Therefore, each person sample in the training set carries three labels: the intra-camera ID, the camera label, and the global ID. Moreover, since the same pedestrian might be captured by multiple cameras, they have different IDs assigned to them depending on which camera they were captured by. Thus, our primary objective is to learn feature representation for individuals across different cameras.

Recently, the CLIP model, trained on large-scale datasets, has demonstrated remarkable proficiency in matching image-text descriptions. Its image encoder captures complex and rich visual features, while the text encoder provides enhanced semantic information. Building upon this, we have developed a learning framework that integrates CLIP with ICS-ReID to discern intra- and inter-camera ID identities, as illustrated in Fig . \ref{fig:overview}. The framework consists of three training steps: intra-camera pre-defined prompt learning, intra-camera learning, and inter-camera learning. Through this three-stage training process, the model can deeply explore pedestrian information from different camera angles and effectively guide the establishment of more accurate cross-camera ID associations.

% First, we utilize the independently annotated labels within each camera and the prompt learning of the text encoder to obtain corresponding implicit textual description features. This provides semantic support for subsequent intra-camera supervised learning and inter-camera unsupervised learning. In the intra-camera learning phase, we construct a hybrid memory bank based on the supervised information within each camera. This bank stores the central features of pedestrians and all corresponding instance features within each camera. We also design an intra-camera discriminative contrastive loss to learn the feature representations of each pedestrian within that camera.
% In the inter-camera learning phase, we use a clustering algorithm to associate pedestrians across cameras, constructing an inter-camera memory to store the centroid features of associated pedestrian IDs. Through inter-camera contrastive learning, we bring closer the representations of the same pedestrian across different cameras. 

% Additionally, we introduce the Gradient Reversal Layer (GRL) \cite{ganin2015unsupervised} method to build a camera classifier, which helps the model learn the feature consistency of pedestrians across different cameras.

\subsection{Intra-camera Pred-defined Labels Prompt Learning}
The current research on ICS ReID commonly adopts a two-stage learning approach, namely intra- and inter-camera stages. In the inter-camera learning phase, pseudo-labels are often assigned to IDs from different camera views using similarity-matching. However, due to variations in perspective, these pseudo-labels tend to be inaccurate, which can hinder the learning process. To address this issue, we integrate text encoder and prompt learning mechanisms from the CLIP framework to generate descriptive textual prompts corresponding to individual identities. This incorporation provides valuable semantic constraints for subsequent inter-camera learning and serves as an adjunct in rectifying pseudo-labels, thereby enhancing the overall recognition performance.

More specifically, we first define a textual prompt based on the predefined labels within each camera, described as "a photo of $[X]_1$$[X]_2$…$[X]_M$ person," where \( M \) represents the number of learnable text tokens. Subsequently, we input the implicit textual prompts and ID images into the CLIP model to optimize the text tokens $[X]$. Through this method, we can obtain textual representations associated with pedestrian IDs within each camera. It is important to note that, in this training phase, we freeze CLIP model's image and text encoder while utilizing image-to-text and text-to-image losses to learn the text tokens:

\begin{equation}
  {\mathcal L_{i2t}} =  - \sum\limits_{c = 1}^C {\frac{1}{{\left| {{P_i}} \right|}}\sum\limits_{p \in {P_i}} {\log \frac{{\exp \left( {s\left( {{f_i^v},{f_p^t}} \right)/\tau } \right)}}{{\sum\nolimits_{k = 1}^B {\exp \left( {s\left( {{f_i^v},{f_k^t}} \right)/\tau } \right)} }}} },
  \label{eq:01}
\end{equation}
\begin{equation}
  {\mathcal L_{t2i}} =  - \sum\limits_{c = 1}^C {\frac{1}{{\left| {{P_i}} \right|}}\sum\limits_{p \in {P_i}} {\log \frac{{\exp \left( {s\left( {{f_p^v},{f_i^t}} \right)/\tau } \right)}}{{\sum\nolimits_{k = 1}^B {\exp \left( {s\left( {{f_k^v},{f_i^t}} \right)/\tau } \right)} }}}} ,
  \label{eq:02}
\end{equation}
where ${P_i} = \{ p|{y_p} = {y_i},p \in \{ 1,2,...,B\} \}$ represents the index set of positive image samples, $s(,)$ represents the cosine similarity between text features and image features, $C$ is the number of cameras, $B$ is the batch size, and $\tau$ denotes the temperature factor. Ultimately, the loss of intra-camera pre-defined label prompt learning is:
\begin{equation}
  {\mathcal L_{prompt}} = {\mathcal L_{i2t}} + {\mathcal L_{t2i}}.
  \label{eq:03}
\end{equation}

By minimizing \( \mathcal L_{prompt} \), we can learn the corresponding implicit textual descriptions for the IDs within each camera. In the subsequent stages, we will utilize these textual descriptions to provide stronger semantic supervision for the model, thereby enhancing its generalization capability.

\subsection{Intra-camera Discriminative Learning}
The primary challenge of ICS ReID lies in annotating IDs within each camera's view and establishing cross-camera ID associations through the analysis of inter-camera characteristics. Intra-camera learning, therefore, can be considered a fully supervised problem within a multi-task framework. 
However, the distribution of sample numbers for each ID within a camera is uneven, with most IDs having only a limited set of training samples. This imbalance can lead to model bias toward learning prominent camera-style features, rather than ID features. Additionally, the lack of cross-camera ID information highlights the significance of intra-camera learning as a preparatory phase for subsequent cross-camera correlation.

Considering these factors, we focus on the centroid features of each pedestrian within each camera, as well as the hard positive and hard negative samples for that pedestrian within the camera, as shown in Fig. \ref{fig:intra}. This compels the model to learn more accurate intra-camera pedestrian features. This approach also emphasizes the differences between pedestrian IDs within the camera view, providing more reliable features for subsequent inter-camera association steps.

\subsubsection{Intra-camera Hybrid Memory Banks Initialization}
Firstly, we initialize the intra-camera hybrid memory bank using an image feature encoder. All image features are assigned to the corresponding camera memory based on different camera IDs. Then, for each image under each camera's pre-defined labels, the centroid features of each pedestrian ID within the camera are stored in the intra-camera centroid memory through averaging, to learn the pedestrian features within each camera. Additionally, the instance features corresponding to each pedestrian ID within each camera are stored in the intra-camera instance memory to learn more discriminative information. The mean feature of the predefined labels of pedestrians within a camera is calculated as follows:
\begin{equation}
   \mu_{c}^{i} = \frac{1}{|N_{c}^{i}|} \sum_{x \in N_{c}^{i}} f({x}),
  \label{eq:04}
\end{equation}
where \(\mu_{c}^{i}\) represents the mean feature of pedestrian ID \(i\) within camera \(c\), \(N_{c}^{i}\) denotes the set of all images belonging to pedestrian ID \(i\) within camera \(c\), and \(f({x})\) represents the features of image \(x\) after being processed by the image encoder. Therefore, the memory bank \(M_{\text{c}}^{intra}\) is initialized with the mean features of the pedestrian IDs, while \(M_{\text{i}}^{intra}\) is initialized with the instance sample features corresponding to those pedestrian IDs. For intra-camera instance memory, we employ a real-time instance feature memory update strategy. In each iteration, we directly replace \(M_{\text{i}}^{intra}\) in the memory with the current mini-batch instantaneous feature \( f_x \):
\begin{equation}
   M_i^{intra}\left[ y \right] \leftarrow f(x).
  \label{eq:05}
\end{equation}

\subsubsection{Optimization}
In each training iteration, the features stored in the aforementioned intra-camera hybrid memory bank are updated using different strategies.

\begin{figure}[t]
  \centering
  \includegraphics[width=0.5\textwidth]{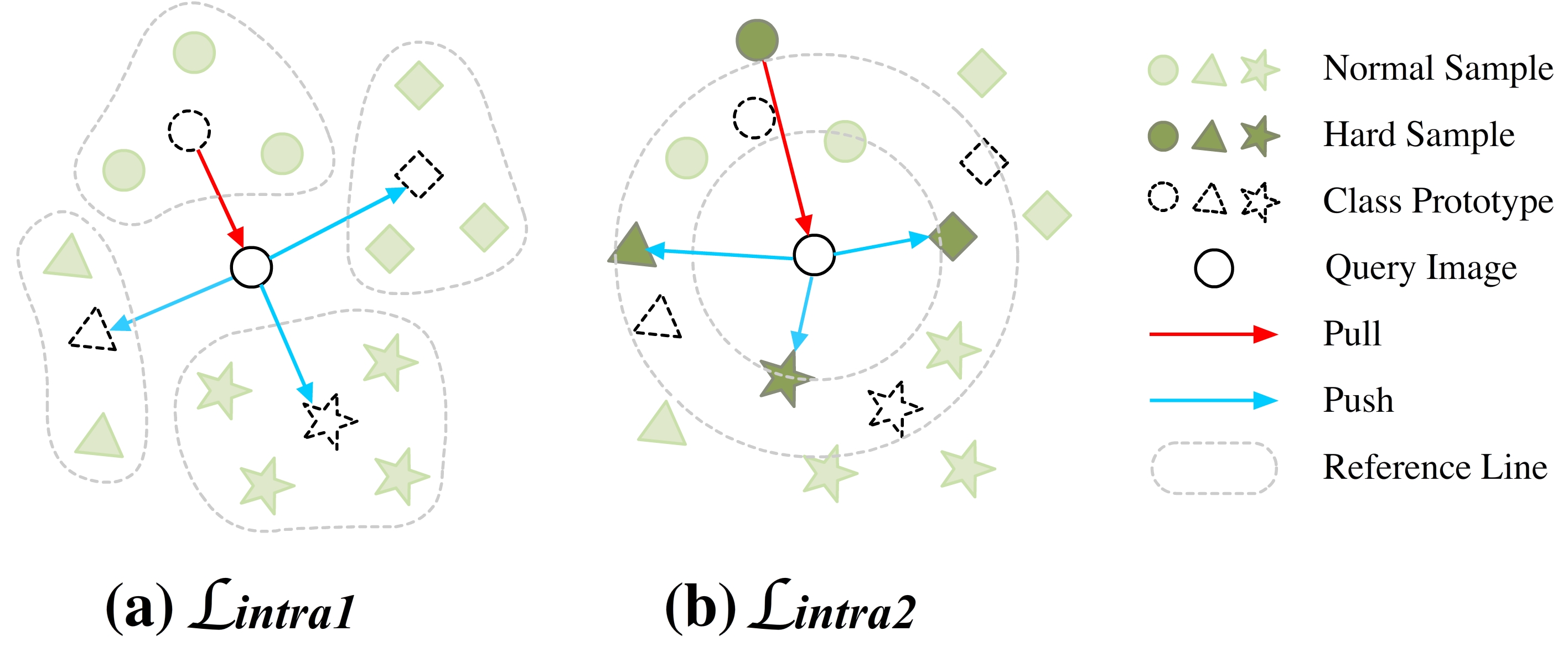}
  \caption{Illustration of $\mathcal L_{intra1}$ and $\mathcal L_{intra2}$. The same color indicates that all samples originate from the same camera, while different shapes represent different pedestrian IDs within the camera.
  }
  \label{fig:intra}
\end{figure}

For the memory bank \(M_{\text{c}}^{intra}\):

\begin{equation}
   \mu_{c}^{i} \leftarrow \alpha \mu_{c}^{i} + (1 - \alpha) \widetilde \mu _c^i,
  \label{eq:06}
\end{equation}
where \(\widetilde \mu _c^i\) denotes the average features of the camera \(c\) insider ID \(i\) in each batch, \(\alpha\) is the momentum updating factor.
This update mechanism ensures that the features in the memory bank consistently reflect the latest training information, thereby enhancing the accuracy and stability of the model in learning intra-camera pedestrian features, as shown in Fig. \ref{fig:intra} (a).
To this end, given a query image feature \(f({x})\), we propose an intra-camera centroid contrastive loss function, which is
formulated as:

\begin{equation}
  \mathcal{L}_{intra1} = - \sum_{c=1}^{C} \log \frac{\exp \left( s(f(x_i), \mu^{+}) / \tau \right)}{\sum_{j=1}^{K_c} \exp \left( s(f(x_i), \mu_{c}^{i}) / \tau \right)},
  \label{eq:07}
\end{equation}
where $\mu$ represents the centroid feature for each ID in the $c$-th intra-camera memory, $K_c$ represents the total number of pedestrian IDs in the camera, and $C$ is the number of cameras. Through the above loss, we can effectively bring the sample closer to the centroid feature of its corresponding ID while pushing it away from other ID centroid features within the same camera. 

However, when faced with challenging samples within the camera, such as similar apparel or shared backgrounds, this approach may result in poor classification of IDs within the camera. Moreover, as the dataset expands and the number of individuals per camera increases, the model's recognition performance may be adversely affected. Therefore, we further enhance inter-class separability and intra-class compactness by merging all instance features under each ID, as shown in Fig. \ref{fig:intra} (b). Specifically, for a query image ${x_i}$, we examine the relationship between the hardest positive sample and the hardest negative sample from other IDs stored in the memory. By calculating loss across different cameras, we reduce the distance between samples and their centroids as well as relevant hard positive samples while increasing distances from other hard negative samples.
\begin{equation}
  {\mathcal L_{intra2}} =  - \sum\limits_{c = 1}^C {\log } \frac{{\exp (s(f({x}),m_{hard}^ + )/\tau )}}{{\sum\nolimits_{j = 1}^{K_c} {\exp (s(f({x}),m_{hard}^ j )/\tau )} }},
  \label{eq:08}
\end{equation}
where $m_{hard}^ +$ represents the hardest positive sample characteristics retained in memory \(M_{\text{i}}^{intra}\), it demonstrates the cosine similarity is the lowest compared to all instance features under this pedestrian ID.
Conversely, $m_{hard}^ j$ is the hardest negative sample feature which shows the highest cosine similarity when compared to all other instance features from different IDs under the camera. 

\subsubsection{Intra-camera Image-Text Alignment}
During the second stage of intra-camera learning, we freeze the text encoder and only train the image encoder. Specifically, for each person ID under each camera, we obtain the corresponding text features by inputting prompts into the text encoder, derived from the first stage. Meanwhile, we input the image \( x \) into the image encoder to obtain the raw features \( {f^v}(x) \). Subsequently, we use the loss \( \mathcal{L}_{i2tce} \) to constrain the image features \( {f^v}(x) \) to be close to the corresponding text features \( {f^t}(y) \), while being distant from the text features of other identities:

\begin{equation}
{\mathcal{L}_{i2tce}^{intra}} = \sum_{c = 1}^C \sum_{i = 1}^{K_c} -{q_z} \log \frac{\exp(s(f^v(x), f^t(y_i)))}{\sum_{z = 1}^{K_c} \exp(s(f^v(x), f^t(y_z)))} ,
\label{eq:09}
\end{equation}
where ${q_z}$ is a smoothed ID label. Notably, our loss function is computed for features within each camera independently, ensuring that these calculations are performed separately for each camera.

\subsubsection{The Loss for Intra-camera Learning}

In summary, the proposed intra-camera loss $\mathcal{L}_{\textit{intra}}$ is composed of the intra-camera discriminative loss and the intra-camera image-text alignment loss, defined as follows:
\begin{equation}
  {\mathcal L_{ICDL}} = \lambda {\mathcal L_{intra_1}} + (1 - \lambda ){\mathcal L_{intra2}},
  \label{eq:10}
\end{equation}
where $\lambda$ is the balancing factor between the two losses.

\begin{equation}
  {\mathcal L_{intra}} = {\mathcal L_{ICDL}} + {\mathcal L_{i2tce}^{intra}}.
  \label{eq:11}
\end{equation}

\subsection{Inter-camera Learning}
Through the aforementioned intra-camera learning, our model effectively identifies each person within the camera's view. However, the abundant learning information of IDs across cameras has yet to be fully utilized. Therefore, in the inter-camera learning process, we have devised an alternating strategy consisting of cross-camera ID association steps and inter-camera contrastive learning steps to facilitate the model's acquisition of view-invariant ID features.
\subsubsection{Inter-camera Association}
The ICS ReID approach differs from fully unsupervised ReID, which relies on clustering algorithms to obtain pseudo-labels directly. In the case of intra-camera IDs, assigning the same pseudo-labels is not feasible. Therefore, we employ an ID association algorithm based on connected components proposed in \cite{wang2021towards}. Specifically, we impose two constraints on the clustering process: 1) under the in-camera supervised condition, positive matches among IDs within each camera should not exist, and 2) a maximum of one positive match is allowed per camera. We then constructed an undirected graph $G = \langle V, E\rangle$ for associations, where the vertex set $V$ represents the accumulated IDs across all cameras, and the edge set $E$ represents a positive pair between IDs $i$ and $j$. The edge $e(i,j)$ is defined as follows:
\begin{equation}
 e(i,j){\rm{ = }}\left\{ \begin{array}{l}
1,~dist(i,j) < T \wedge c(i) \ne c(j)\\
\begin{array}{*{20}{c}}
{}& \wedge 
\end{array}i \in {N_1}(j,c(i)) \wedge j \in {N_1}(i,c(j));\\
0,~otherwise.
\end{array} \right.
  \label{eq:12}
\end{equation}
where $dist(i,j)$ represents the distance between the centroid features of the $i$-th and $j$-th ID IDs, $c(i)$ indicates the camera to which the IDs belong, and $T$ is the distance threshold. ${N_1}(j,c(i))$ designates the nearest neighbor of the $j$-th ID with the $i$-th ID under the $c(i)$ camera. Using the conditions defined above, a sparsely connected graph is constructed. Then, based on similarity, we identify all connected components and assign inter-camera pseudo-labels to IDs.

\subsubsection{Inter-camera Memory Banks Initialization}
Based on the successful application of contrastive learning in the field of person re-identification \cite{dai2022cluster,zhang2022implicit,wang2022offline}, we employ a prototypical contrastive learning paradigm for inter-camera learning. First, upon completion of intra-camera learning, we generate pseudo-labels for IDs using the aforementioned inter-camera association algorithm. Next, we compute the mean features of these samples based on their corresponding pseudo-labels and directly initialize the inter-camera memory bank. This approach provides a stable starting point for prototypical contrastive learning. Consequently, our inter-camera memory stores the mean features of the associated IDs across different cameras, facilitating the learning of a person's appearance characteristics under varying camera conditions. The memory features are updated using online batch features in a moving average manner, as described by the following formula:
\begin{equation}
M[{y}] \leftarrow \alpha M[{y}] + (1 - \alpha ){f_{x}}.
  \label{eq:13}
\end{equation}

\subsubsection{Optimization}
To further learn the prototype features of IDs under different cameras, the inter-camera prototypical contrastive loss is defined as follows:
\begin{equation}
{\mathcal L_{IPCL}} =  - \log \frac{{\exp (s(f({x}),M[{y}])/\tau )}}{{\sum\nolimits_{i = 1}^Z {\exp (s(f({x}),M[j])/\tau )} }},
  \label{eq:14}
\end{equation}
where $Z$ represents the number of IDs associated in each epoch of inter-camera correlation.

\subsubsection{Inter-camera Image-Text Alignment}
Considering the significant variations in illumination and background, IDs across cameras often exhibit notable feature differences, leading to noisy inter-camera association labels. Hence, we combine the text description learned in the first stage with the inter-camera prototypical contrastive learning, leveraging additional semantic supervision information to assist the model in improving the accuracy of inter-camera ID correlation and learning the prototype features of IDs across cameras. Specifically, we define an image-to-text contrastive loss function:
\begin{equation}
{\mathcal{L}_{i2tce}^{{inter}}} = \sum_{i = 1}^{Z} -{q_z} \log \frac{\exp(s(f^v(x), f^t(y_i)))}{\sum_{z = 1}^{Z} \exp(s(f^v(x), f^t(y_z)))} ,
\label{eq:15}
\end{equation}

\subsubsection{The Loss for Inter-camera Learning}
The total loss in inter-camera learning can be summarized as follows:
\begin{equation}
{\mathcal L_{inter}} = {\mathcal L_{IPCL}} + {\mathcal L_{i2tce}^{inter}}.
  \label{eq:16}
\end{equation}

\begin{figure*}[t]
  \centering
    \includegraphics[width=1.0\linewidth]{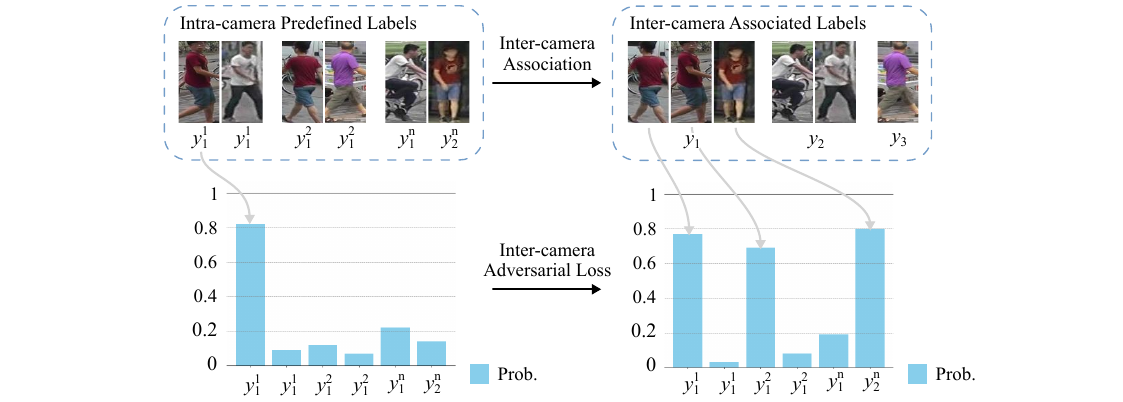}
  \caption{The probability distributions of intra-camera samples processed by the global classifier in Market-1501 dataset are as follows: The left figure illustrates that, after a certain number of training epochs, the samples with true intra-camera labels exhibit a distinct probability distribution with a sharp peak, indicating that the classifier effectively distinguishes different individuals across different cameras. The right figure shows that, after initiating inter-camera adversarial learning, inter-camera association labels are obtained through an inter-camera association algorithm. Samples with the same pseudo-label across different cameras are treated as positive examples, which enhances the probability distribution of the same person across cameras in the global classifier, resulting in multiple peaks.}
  \label{fig:ICAL}
\end{figure*}
 %  The process of training a GI classifier and the
 % adversarial relationship between the model and the classifier.
 % The dashed line represents detaching the features before
 % inputting them into the classifier during the training phase.
 % Subsequently, the classifier outputs q and uses the global
 % ID is the positive class to train the classifier, enabling
 % it to distinguish images of the same person captured by
 % different cameras. During the adversarial phase, the features
 % are directly inputted into the classifier and get q. Then
 % all classes that have the same pseudo-label as the query are
 % considered positive examples for training. This allows the
 % model to deceive the classifier and establish an adversarial
 % relationship. Where + indicates positive class and- indicates
 % negative class.

\subsection{Inter-camera Adversarial Learning} 
Our model can recognize pedestrian identities across different cameras through the previously described intra- and inter-camera contrastive learning. However, during inter-camera learning, the variance in pedestrian feature distributions between cameras introduces label noise, and the value of predefined intra-camera label information is not fully exploited. To address this issue, we propose Inter-Camera Adversarial Loss (ICAL). ICAL penalizes the model's prediction capabilities across different cameras, forcing the backbone network to extract camera-agnostic features. To achieve this, we introduce a new global ID classifier based on camera data, appended to the network, as shown in Fig . \ref{fig:overview}.
Each training iteration consists of two optimization steps:
\subsubsection{Training the Inter-Camera Global ID Classifier} First, we establish a classifier ${C^C}( \cdot )$, where each class corresponds to a global ID. We optimize this global ID classifier by minimizing the classification loss ${\mathcal{ L}_{GID}}$, defined as the cross-entropy loss between the predicted pedestrian 
${C^C}(f({x_i}))$ and the global label 
$y_i^G$. We perform $L2$-normalization on the model's output features $f({x_i})$ and denote the $L2$-normalized weights of the $j$-th global ID classifier as ${\varphi _j}$. We detach these weights before inputting them into the global ID classifier to ensure that the classifier's training does not influence the model itself. Consequently, ${\mathcal{ L}_{GID}}$ can be expressed as:

\begin{equation}
{\mathcal{L}_{GID}} =  - \sum\limits_{i = 1}^N {\log \frac{{\exp \left( {f({x_i}) \cdot {\varphi _{{y_i^G}}}/\tau } \right)}}{{\sum\limits_{j = 1}^{{N_G}} {\exp \left( {f({x_i}) \cdot {\varphi _j}/\tau } \right)} }}},
\label{eq:17}
\end{equation}
where $N$ is the batch size, $N_G$ is the number of global ID classes in the training set, and $\tau$ is a temperature parameter.
By using global IDs as labels during training, our classifier can distinguish pedestrians across different cameras, effectively predicting which specific camera a pedestrian image feature originates from.

\begin{table*}[t]
  \caption{Details of each dataset. In training set, $\#Camera$, $\#ID$, and $\#Image$ are the number of cameras, IDs, and images, respectively. Under the ICS setting, $\# I{D_{ICS}}$ represents the number of accumulated IDs. In testing set, the number of IDs and images in gallery and query sets are also listed.}
  \label{tab:dataset}
  \centering
\begin{tabular}{c|c|ccc|cc|cc}
\hline
\multirow{2}{*}{Dataset} & \multirow{2}{*}{$\# Camera$} & \multicolumn{3}{c|}{Training Set}                                & \multicolumn{2}{c|}{Gallery Set} & \multicolumn{2}{c}{Query Set}   \\ \cline{3-9} 
                         &                         & \multicolumn{1}{c|}{$\#ID$}  & \multicolumn{1}{c|}{$\#Image$} & \multicolumn{1}{c|}{\textbf{$\# I{D_{ICS}}$}} & \multicolumn{1}{c|}{$\#ID$}  & $\#Image$ & \multicolumn{1}{c|}{$\#ID$}  & $\#Image$ \\ \hline
Market1501               & 6                       & \multicolumn{1}{c|}{751}   & \multicolumn{1}{c|}{12,936} & 3,262      & \multicolumn{1}{c|}{751}    &15,913       & \multicolumn{1}{c|}{750}   &3,368      \\
DukeMTMC-ReID            & 8                       & \multicolumn{1}{c|}{702}    & \multicolumn{1}{c|}{16,522}       &2,196       & \multicolumn{1}{c|}{1,110}    &17,661      & \multicolumn{1}{c|}{702}   & 2,228      \\
MSMT17                   & 15                      & \multicolumn{1}{c|}{1,041} & \multicolumn{1}{c|}{32,621}     & 4,821     & \multicolumn{1}{c|}{3,060}    &82,161      & \multicolumn{1}{c|}{3,060}   &11,659       \\ \hline
\end{tabular}
\end{table*}

\subsubsection{Learning Camera-irrelevant Features}
In the second step, we fix the parameters of the global ID classifier and compel the network to learn camera-irrelevant features. To achieve this, we penalize the model's prediction capability regarding global IDs. Specifically, while the classifier mentioned above can distinguish different pedestrians across cameras using global IDs, our goal is to train the global ID classifier to not distinguish the same identity across different cameras. Therefore, we introduce an Inter-Camera Adversarial Loss (ICAL), a multi-positive class classification loss where all categories belonging to the same identity but different cameras are considered positive classes, as shown in Fig. \ref{fig:ICAL}.%For instance, for a given sample $x_i$, all categories of its identity class $y_i$ are defined as its positive classes. 
Notably, the same pedestrians across different cameras are identified using pseudo-labels obtained via the above inter-camera association algorithm. The ICAL is formulated as follows:

\begin{equation}
\begin{aligned}
\mathcal{L}_{ICAL} = & - \sum\limits_{i = 1}^N \sum\limits_{g = 1}^{N_G} q(g) \\
& \log \left( \frac{\exp \left( \frac{f(x_i) \cdot \varphi_g}{\tau} \right)}{\exp \left( \frac{f(x_i) \cdot \varphi_g}{\tau} \right) + \sum\limits_{j \in G_i^-} \exp \left( \frac{f(x_i) \cdot \varphi_j}{\tau} \right)} \right),
\end{aligned}
\label{eq:18}
\end{equation}

where $G_i^-$ represents the set of global IDs of the negative centroids of the query. $q(g)$ is the cross-entropy loss weight for the $g$-th global ID category. 

Given that the same pedestrians across different cameras are identified using pseudo-labels obtained via the inter-camera association algorithm, which contains noise compared to pre-defined intra-camera labels, to enhance the model's inter-camera pedestrian recognition capability without significantly compromising intra-camera accuracy, we define $q(g)$ as:

\begin{equation}
q(g) = \left\{ 
\begin{array}{ll}
1 - \epsilon + \frac{\epsilon}{G}, & \text{if } g = y_i^G \\
\frac{\epsilon}{G}, & \text{if } g \ne y_i^G \text{ and } g \in G_i^+ \\
0, & \text{if } g \in G_i^-
\end{array}
\right.
\label{eq:19}
\end{equation}
where $G_i^+$ represents the set of global IDs of the positive centroids of the query, $G$ is the number of elements in $G_i^+$, and $G_i^-$ represents the set of global IDs of the negative centroids of the query. $\epsilon$ is a hyperparameter with a range of
$0 < \epsilon  \le 1$.

Importantly, our goal is to optimize both ICAL and the inter-camera loss concurrently. ${\mathcal{L}_{inter}}$ and ${\mathcal{L}_{ICAL}}$ are correlated in learning camera-irrelevant features. When using ${\mathcal{L}_{inter}}$ alone, the model tends to learn simple samples (highly similar pedestrian features across different cameras) in the early epochs of optimization and gradually distinguishes harder samples (pedestrian features with low similarity due to factors like pose, lighting, and background changes). ${\mathcal{L}_{ICAL}}$ aims to narrow the feature gap for the same identity across different cameras with pseudo-labels, which is similar to the goal of ${\mathcal{L}_{inter}}$. To avoid local optima caused by directly minimizing ${\mathcal{L}_{ICAL}}$ from the beginning, we execute inter-camera adversarial learning after a certain number of epochs. Consequently, our model improves inter-camera pedestrian recognition capability and effectively learns camera-irrelevant features without significantly compromising intra-camera recognition accuracy.

\subsection{A Summary of the Objective Function}
According to the above description, we name the CLIP-based camera-agnostic feature learning framework CCAFL. The overall loss function of CCAFL is:
\begin{equation}
\mathcal{L}_{CCAFL} = 
\begin{cases} 
\mathcal{L}_{intra} + \mathcal{L}_{GID}, & \text{if } epoch \le E_{intra}, \\
\mathcal{L}_{inter} + \mathcal{L}_{GID}, & \text{if } E_{intra} < epoch, \\
\mathcal{L}_{inter} + \mathcal{L}_{GID} + \mathcal{L}_{ICAL}, &  \text{if } E_{adv} \le epoch.
\end{cases}
\label{eq:20}
\end{equation}
where $E_{intra}$ and $E_{adv}$ denote the number of epochs for intra-camera and inter-camera adversarial learning, respectively. Through the proposed learning process, the CCAFL algorithm not only effectively utilizes the textual information generated by CLIP to provide semantic supervision for subsequent learning but also leverages pre-defined labels within each camera for supervised intra-camera learning and adversarial inter-camera learning. This approach enables the model to learn camera-agnostic features, thereby enhancing the quality of inter-camera clustering.

\section{Experiments}
\subsection{Datasets and Evaluation Metrics}
% Our method is validated on three large-scale person ReID datasets: Market-1501 \cite{zheng2015scalable}, DukeMTMC-ReID \cite{ristani2016performance}, and MSMT17 \cite{wei2018person}. According to the ICS setting, we re-annotate the individuals within each camera in the training set and add accumulation labels. The specific circumstances of the dataset under the ICS setup are summarized in Table \ref{tab:dataset}.

Our method is validated on three large-scale person re-identification (ReID) datasets: Market-1501 \cite{zheng2015scalable}, DukeMTMC-ReID \cite{ristani2016performance}, and MSMT17 \cite{wei2018person}. Following the ICS setting, we re-annotate individuals within each camera in the training set and add accumulation labels. Table \ref{tab:dataset} summarizes the specifics of the datasets under the ICS setup, including the number of cameras, IDs, and images in the training, gallery, and query sets. Additionally, we provide the accumulated total identity number under intra-camera supervision ($\# I{D_{ICS}}$).

% \textit{Market1501} contains 36,036 images captured by 6 cameras, involving 1,501 IDs. The training set consists of 12,936 images from 751 identities, with the average of 17.23 images per person. While for ICS setting, the average number of images per camera per person is 3.97, and the training data involves 3,262 IDs.

% \textit{DukeMTMC-ReID} contains 36,411 images captured by 8 cameras, involving 1,404 IDs. The training set contains 16,522 images from 702 identities, while these images involve 2,196 identities in the ICS setting, and the average number of images per camera per person is 7.52. 

% \textit{MSMT17} contains 126,441 images of 4,101 identities captured by 15 cameras, including 32,621 images of 1,041 IDs in the training set, whereas, for the ICS setting, the training data includes 4,821 identities.

In terms of evaluation metrics, we adopt cumulative matching characteristics (CMC) \cite{gray2007evaluating} including Rank-1, Rank-5, and Rank-10 as well as mean Average Precision (mAP).

\subsection{Implementation Details}
We adopt the ResNet50 model pre-trained on CLIP as our feature extractor. All input images are resized to 256×128 and subjected to data augmentation techniques such as random flipping, cropping, and erasing \cite{zhong2020random}. During the prompt learning phase, we learn tokens represented as $[X]_1$$[X]_2$…$[X]_M$. We utilize the Adam optimizer \cite{kingma2014adam} with a learning rate of 0.00035, which is adjusted using a cosine annealing policy. Our training batch size is 64, and the training process lasts for 60 epochs. For the subsequent training phase, we set the batch size to 128 and employ the PK sample \cite{hermans2017defense}. Within each mini-batch, we sample images based on the labels of each camera. We randomly select 16 IDs, with each ID having 8 images. In this stage, we use the Adam optimizer with a learning rate of 0.00035. The model is warmed up for 10 epochs using a linear growth strategy, and the training phase lasts for 80 epochs. During the initial 5 epochs, we perform only intra-camera learning steps. Subsequently, we alternate between inter-camera association and inter-camera learning steps every epoch. The memory updating rates in Eq. \ref{eq:06} and Eq. \ref{eq:12} are set to 0.1, and the balance weight in Eq. \ref{eq:10} is set to 0.8. The temperature parameter $\tau$ in Eq. \ref{eq:07}, Eq. \ref{eq:08}, and Eq. \ref{eq:13} is set to 0.05. The hyperparameter $\epsilon$ in Eq. \ref{eq:19} is set to 0.8 for all datasets. The distance threshold $T$ in Eq. \ref{eq:11} is set to 1.7 for Market-1501 and 1.5 for both DukeMTMC-ReID and MSMT17. All experiments are conducted using PyTorch on four NVIDIA Tesla A100 GPUs.

\begin{table*}[t]
  \caption{Comparison with state-of-the-art methods on the Market-1501, DukeMTMC-ReID, and MSMT17 datasets, including the best results among intra-camera supervised (ICS) ReID methods as well as state-of-the-art fully supervised (SUP) ReID methods and purely unsupervised (USL) ReID methods. The best results in ICS methods are indicated in BOLD.}
  \label{tab:sota}
\centering
   %\scalebox{0.8}{%%
   \resizebox{\textwidth}{!}{
\begin{tabular}{lcc|cccc|cccc|cccc}
\hline
\multicolumn{3}{c|}{Settings}                                                                          & \multicolumn{4}{c|}{Market-1501}                              & \multicolumn{4}{c|}{DukeMTMC-ReID}                            & \multicolumn{4}{c}{MSMT17}                                    \\ \hline
\multicolumn{1}{c|}{Type}                       & \multicolumn{1}{c|}{Methods}         & Reference     & mAP           & R1            & R5            & R10           & mAP           & R1            & R5            & R10           & mAP           & R1            & R5            & R10           \\ \hline
\multicolumn{1}{l|}{\multirow{6}{*}{SUP-ReID}}  & \multicolumn{1}{c|}{PCB \cite{sun2019learning}}             & ECCV18        & 81.6          & 93.8          & -             & -             & 69.2          & 83.3          & -             & -             & 40.4          & 68.2          & -             & -             \\
\multicolumn{1}{l|}{}                           & \multicolumn{1}{c|}{BoT \cite{luo2019bag}}             & CVPRW19       & 85.9          & 94.5          & -             & -             & 76.4          & 86.4          & -             & -             & 45.1          & 63.4          & -             & -             \\
\multicolumn{1}{l|}{}                           & \multicolumn{1}{c|}{ABD-Net \cite{chen2019abd}}         & ICCV19        & 88.3          & 95.6          & -             & -             & 78.6          & 89.0          & -             & -             & 60.8          & 82.3          & -             & -             \\
\multicolumn{1}{l|}{}                           & \multicolumn{1}{c|}{AGW \cite{ye2021deep}}             & TPAMI21       & 87.8          & 95.1          & -             & -             & 79.6          & 89.0          & -             & -             & 49.3          & 68.3          & -             & -             \\
\multicolumn{1}{l|}{}                           & \multicolumn{1}{c|}{LTReID \cite{9786762}}          & TMM22         & 89.0          & 95.9          &               &               & 80.4          & 90.5          &               &               & 58.6          & 81.0          &               &               \\
\multicolumn{1}{l|}{}                           & \multicolumn{1}{c|}{CLIP-ReID \cite{li2023clip}}       & AAAI23        & 89.8          & 95.7          & -             & -             & 80.7          & 90.0          & -             & -             & 63.0          & 84.4          & -             & -             \\ \hline
\multicolumn{1}{l|}{\multirow{15}{*}{USL-ReID}} & \multicolumn{1}{c|}{MMT \cite{ge2020mutual}}             & ICLR20        & 71.2          & 87.7          & 94.9          & 96.9          & 65.1          & 78.0          & 88.8          & 92.5          & 23.3          & 50.1          & 63.9          & 69.8          \\
\multicolumn{1}{l|}{}                           & \multicolumn{1}{c|}{SPCL \cite{ge2020self}}            & NeurIPS20     & 73.1          & 88.1          & 95.1          & 97.0          & 64.4          & 80.0          & 89.0          & 91.6          & 19.1          & 42.3          & 55.6          & 61.2          \\
\multicolumn{1}{l|}{}                           & \multicolumn{1}{c|}{RLCC \cite{9577618}}            & CVPR21        & 77.7          & 90.8          & 96.3          & 97.5          & 69.2          & 83.2          & 91.6          & 93.8          & 27.9          & 56.5          & 68.4          & 73.1          \\
\multicolumn{1}{l|}{}                           & \multicolumn{1}{c|}{IICS \cite{xuan2021intra}}            & CVPR21        & 72.9          & 89.5          & 95.2          & 97.0          & 64.4          & 80.0          & 89.0          & 91.6          & 26.9          & 56.4          & 68.8          & 73.4          \\
\multicolumn{1}{l|}{}                           & \multicolumn{1}{c|}{ICE \cite{chen2021ice}}             & ICCV21        & 79.5          & 92.0          & 97.0          & 98.1          & 67.2          & 81.3          & 90.1          & 93.0          & 29.8          & 59.0          & 71.7          & 77.0          \\
\multicolumn{1}{l|}{}                           & \multicolumn{1}{c|}{CAP \cite{wang2021camera}}             & AAAI21        & 79.2          & 91.4          & 96.3          & 97.7          & 67.3          & 81.1          & 89.3          & 91.8          & 36.9          & 67.4          & 78.0          & 81.4          \\
\multicolumn{1}{l|}{}                           & \multicolumn{1}{c|}{HDCPD \cite{cheng2022hybrid}}           & TIP21         & 84.5          & 93.5          & 97.6          & 98.6          & 73.5          & 85.4          & 92.2          & 94.5          & 20.7          & 43.8          & 55.1          & 60.1          \\
\multicolumn{1}{l|}{}                           & \multicolumn{1}{c|}{GroupSampling \cite{han2022rethinking}}   & TIP22         & 79.2          & 92.3          & 96.6          & 97.8          & 69.1          & 82.7          & 91.1          & 93.5          & 24.6          & 56.2          & 67.3          & 71.5          \\
\multicolumn{1}{l|}{}                           & \multicolumn{1}{c|}{ClusterContrast \cite{dai2022cluster}} & ACCV22        & 83.0          & 92.9          & 97.2          & 98.0          & 73.6          & 85.5          & 92.2          & 94.6          & 31.2          & 61.5          & 71.8          & 76.7          \\
\multicolumn{1}{l|}{}                           & \multicolumn{1}{c|}{PPLR \cite{cho2022part}}            & CVPR22        & 84.4          & 94.3          & 97.8          & 98.6          & -             & -             & -             & -             & 42.2          & 73.3          & 83.5          & 86.5          \\
\multicolumn{1}{l|}{}                           & \multicolumn{1}{c|}{ISE \cite{zhang2022implicit}}             & CVPR22        & 85.3          & 94.3          & 98.0          & 98.8          & -             & -             & -             & -             & 37.0          & 67.6          & 77.5          & 81.0          \\
\multicolumn{1}{l|}{}                           & \multicolumn{1}{c|}{O2CAP \cite{wang2022offline}}           & TIP22         & 82.7          & 92.5          & 96.9          & 98.0          & 71.2          & 83.9          & 91.3          & 93.4          & 42.4          & 72.0          & 81.9          & 85.4          \\
\multicolumn{1}{l|}{}                           & \multicolumn{1}{c|}{CCL \cite{zhang2023camera}}           & TCSVT23         & 85.3          & 94.1          & -          & -          & -          & -          & -          & -          & 41.8          & 71.4          & -          & -          \\
\multicolumn{1}{l|}{}                           & \multicolumn{1}{c|}{DCMIP \cite{zou2023discrepant}}           & ICCV23        & 86.7          & 94.7          & 98.0          & 98.8          & -             & -             & -             & -             & 40.9          & 69.3          & 79.7          & 83.6          \\
\multicolumn{1}{l|}{}                           & \multicolumn{1}{c|}{RTMem \cite{10102757}}           & TIP23         & 86.5          & 94.3          & 97.9          & 98.5          & -             & -             & -             & -             & 38.5          & 63.3          & 75.4          & 79.6          \\
\multicolumn{1}{l|}{}                           & \multicolumn{1}{c|}{LP \cite{10137428}}              & TIP23         & 85.8          & 94.5          & 97.8          & 98.7          & 76.2          & 86.7          & 93.0          & 94.3          & 39.5          & 67.9          & 78.0          & 81.6          \\
\multicolumn{1}{l|}{}                           & \multicolumn{1}{c|}{DHCCN \cite{10453238}}              & TCSVT24         & 85.6          & 94.1          & -          & -          & 73.4          & 83.8          & -          & -          & 36.4          & 65.9          & -          & -          \\ \hline
\multicolumn{1}{l|}{\multirow{10}{*}{ICS-ReID}} & \multicolumn{1}{c|}{UGA \cite{wu2019unsupervised}}             & ICCV19        & 70.3          & 87.2          & -             & -             & 70.3          & 75.0          & -             & -             & 21.7          & 50.2          & -             & -             \\
\multicolumn{1}{l|}{}                           & \multicolumn{1}{c|}{PCSL \cite{qi2020progressive}}            & TCSVT20       & 69.4          & 87.0          & 94.8          & 96.6          & 53.5          & 71.7          & 84.7          & 88.2          & 20.7          & 48.3          & 62.8          & 68.6          \\
\multicolumn{1}{l|}{}                           & \multicolumn{1}{c|}{ACAN \cite{qi2021adversarial}}            & TCSVT21       & 50.6          & 73.3          & 87.6          & 91.8          & 45.1          & 67.6          & 81.2          & 85.2          & 12.6          & 33.0          & 48.0          & 54.7          \\
\multicolumn{1}{l|}{}                           & \multicolumn{1}{c|}{MATE \cite{zhu2021intra}}            & IJCV21        & 71.1          & 88.7          & -             & 97.1          & 56.6          & 76.9          & -             & 89.6          & 19.1          & 46.0          & -             & 65.3          \\
\multicolumn{1}{l|}{}                           & \multicolumn{1}{c|}{Precise-ICS \cite{wang2021towards}}     & WACV21        & 83.6          & 93.1          & 97.8          & 98.6          & 72.0          & 83.6          & 92.6          & 94.7          & 31.3          & 57.7          & 71.1          & 76.3          \\
\multicolumn{1}{l|}{}                           & \multicolumn{1}{c|}{PIRID \cite{wang2022prototype}}           & ICASSP22      & 79.6          & 91.0          & 96.7          & 97.9          & 65.4          & 79.6          & 88.6          & 91.4          & 34.9          & 60.6          & 73.8          & 79.3          \\
\multicolumn{1}{l|}{}                           & \multicolumn{1}{c|}{CDL \cite{peng2022consistent}}             & TMM22         & 84.6          & 94.0          & 97.8          & 98.6          & -             & -             & -             & -             & 48.0          & 76.3          & 86.2          & 89.0          \\
\multicolumn{1}{l|}{}                           & \multicolumn{1}{c|}{DCL \cite{hu2022decoupled}}             & ICPR22        & 86.2          & 95.1          & 98.0          & 98.8          & -             & -             & -             & -             & 45.1          & 74.5          & 84.5          & 87.6          \\
\multicolumn{1}{l|}{}                           & \multicolumn{1}{c|}{CMT \cite{10534060}}             & TCSVT24       & 88.9          & 95.8          & 98.4          & 99.0          & 78.6          & 89.5          & 95.1          & 96.6          & 51.3          & 78.5          & 87.3          & 90.2          \\ \cline{2-15} 
\multicolumn{1}{l|}{}                           & \multicolumn{1}{c|}{\textbf{CCAFL}}  & \textbf{Ours} & \textbf{90.1} & \textbf{96.1} & \textbf{98.6} & \textbf{99.1} & \textbf{81.5} & \textbf{90.8} & \textbf{95.5} & \textbf{97.3} & \textbf{58.9} & \textbf{82.9} & \textbf{90.7} & \textbf{92.8} \\ \hline
\end{tabular}
}
\end{table*}

\begin{table}[t]
  \caption{COMPARISON WITH STATE-OF-THE-ART UNSUPERVISED DOMAIN ADAPTATION METHODS ON THE Market-1501 AND MSMT17 DATASETS. SINCE THEIR PERFORMANCE IS RELATED TO THE AUXILIARY DATASET, WE REPORT THEIR BEST RESULTS. THE BEST RESULTS is INDICATED IN bold values.
  \label{tab:sotauda}}
\centering
 \resizebox{0.45\textwidth}{!}{%
\begin{tabular}{cccccc}
\hline
\multicolumn{1}{c|}{\multirow{2}{*}{Methods}} & \multicolumn{1}{c|}{\multirow{2}{*}{Reference}} & \multicolumn{2}{c|}{Market-1501}                   & \multicolumn{2}{c}{MSMT17}    \\ \cline{3-6} 
\multicolumn{1}{c|}{}                         & \multicolumn{1}{c|}{}                           & mAP            & \multicolumn{1}{c|}{R1}           &  mAP          &  R1           \\ \hline
\multicolumn{6}{l}{\textbf{Unsupervised Domain Adaptation Methods}}                                                                                                                  \\ \hline
\multicolumn{1}{c|}{MMCL \cite{Wang_2020_CVPR}}        & \multicolumn{1}{c|}{CVPR20}          & 60.4          & \multicolumn{1}{c|}{84.4}          &  16.2         &  43.6         \\
\multicolumn{1}{c|}{SPCL \cite{ge2020self}}                     & \multicolumn{1}{c|}{NeurIPS20}                  &  76.7         & \multicolumn{1}{c|}{90.3}          &   26.2       &    53.1        \\
\multicolumn{1}{c|}{RESL \cite{li2022reliability}}                     & \multicolumn{1}{c|}{CVPR20}                     &  83.1        & \multicolumn{1}{c|}{ 93.2}          &      33.6      & 64.8         \\
\multicolumn{1}{c|}{GLT \cite{zheng2021group}}                      & \multicolumn{1}{c|}{CVPR21}                     &79.3          & \multicolumn{1}{c|}{ 90.7}          &  26.5          & 56.6         \\
\multicolumn{1}{c|}{HCD \cite{zheng2021online}}                      & \multicolumn{1}{c|}{ICCV21}                     & 80.2         & \multicolumn{1}{c|}{ 91.4}          &  28.4          & 54.9         \\
\multicolumn{1}{c|}{IDM \cite{dai2021idm}}                      & \multicolumn{1}{c|}{ICCV21}                     &82.8          & \multicolumn{1}{c|}{93.2 }          & 33.5         &  61.3          \\
\multicolumn{1}{c|}{MCRN \cite{wu2022multi}}                     & \multicolumn{1}{c|}{AAAI22}                     & 83.8         & \multicolumn{1}{c|}{93.8 }          &  35.7         & 67.5          \\
\multicolumn{1}{c|}{CaCL \cite{lee2023camera}}                     & \multicolumn{1}{c|}{ICCV23}                     & 84.7          & \multicolumn{1}{c|}{93.8}          &   36.5         &66.6          \\ \hline
\multicolumn{6}{l}{\textbf{Intra-camera Supervised Methods}}                                                                                                                         \\ \hline
\multicolumn{1}{c|}{\textbf{CCAFL}}                 & \multicolumn{1}{c|}{\textbf{Ours}}                       & \textbf{90.1} & \multicolumn{1}{c|}{\textbf{96.1}} & \textbf{58.9} & \textbf{82.9} \\ \hline
\end{tabular}
}
\end{table}

\begin{table*}[tb]
\caption{An effectiveness analysis of different components is conducted on the MSMT17 dataset under all searching configurations.}
\centering
   %\scalebox{1.0}{%%
    \resizebox{0.85\textwidth}{!}{%
\label{tab:xrsy}
\begin{tabular}{c|cccc|cc|cc}
\hline
      & \multicolumn{4}{c|}{Components}    & \multicolumn{2}{c|}{Market-1501} & \multicolumn{2}{c}{MSMT17} \\ \hline
Order & Baseline & $\mathcal L_{ICDL}$ & $\mathcal L_{i2tce}$ & $\mathcal L_{ICAL}$ & mAP              & R1            & mAP           & R1         \\ \hline
0     & \checkmark         &        &        &       &  85.4          &  93.5          &  45.1        &  73.3       \\
1     & \checkmark         &        &        &       &  86.4          &   94.5         &  47.6       & 76.4         \\
2     & \checkmark         &\checkmark       &        &       &     87.7        &   94.7         &  50.2       & 78.4         \\
3     &\checkmark         &        & \checkmark       &       &   88.5         &   95.3         & 52.3        &   79.5      \\
4     &\checkmark         & \checkmark       &\checkmark      &  & 89.2            &  95.5          &  54.6        &80.0         \\
5     & \checkmark         &        &        & \checkmark      &88.2          &  95.4         &  53.5     &   80.7      \\
6     & \checkmark         & \checkmark       &        & \checkmark       & 88.4            &  95.0          &  56.1        &  81.5          \\
7     & \checkmark         & \checkmark       & \checkmark       & \checkmark      &90.1      &  96.1   & 58.9       &  82.9       \\ \hline
\end{tabular}
}
\end{table*}

\subsection{Comparison with State-of-the-Art Methods}
In this section, we compare our proposed CCAFL method with all state-of-the-art ReID methods, The comparison results are summarized in Table \ref{tab:sota} and Table \ref{tab:sotauda}.

\subsubsection{Comparison With Intra-camera Supervised Methods} We compare our proposed CLIP-ICS method with recent ICS ReID methods, including UGA\cite{wu2019unsupervised}, ACAN\cite{qi2021adversarial}, MATE\cite{zhu2021intra}, Precise-ICS\cite{wang2021towards}, PIPRID\cite{wang2022prototype}, CDL\cite{peng2022consistent}, DCL\cite{hu2022decoupled} and CMT\cite{10534060}. Summarizing the comparison results in Table \ref{tab:sota}, we observe that our method outperforms existing ICS methods. Specifically, CCAL surpasses the state-of-the-art CMT method by 1.2\% and 2.9\% in mAP on Market-1501 and DukeMTMC-ReID datasets, respectively, and achieves a 7.6\% improvement on the highly challenging MSMT17 dataset. These results demonstrate the superior performance and generalization capabilities of our method. %By leveraging the powerful semantic processing capabilities of CLIP, we have utilized textual descriptions as an auxiliary tool for supervised learning, significantly enhancing the learning outcomes.

\subsubsection{Comparison With Fully Supervised Methods}
We additionally provide comparisons with four representative fully supervised methods for reference, including PCB \cite{sun2019learning}, BoT \cite{luo2019bag}, ABD-Net \cite{chen2019abd}, AGW \cite{ye2021deep}, LTReID \cite{9786762}, and CLIP-ReID \cite{li2023clip}. As shown in Table \ref{tab:sota}, our approach even surpasses the performance of well-known fully supervised methods such as PCB \cite{sun2019learning}, BoT \cite{luo2019bag} and CLIP-ReID \cite{li2023clip} on Market1501 and DukeMTMC-ReID datasets. Moreover, on the relatively complex MSMT17 dataset, our ICS ReID method achieves performance comparable to CLIP-ReID in terms of Rank-1. These findings demonstrate that our CCAFL method reduces annotation costs while achieving high-performance levels.
 
\subsubsection{Comparison With Purely Unsupervised Methods}
Additionally, our method is comprehensively compared with current state-of-the-art unsupervised methods. Specifically, compared to the state-of-the-art O2CAP method \cite{wang2022offline}, our method outperforms it on MSMT17 dataset by 9.6\% in terms of Rank-1 accuracy and 16.5\% in terms of mAP. Experimental results indicate that in purely unsupervised tasks, the significant variations of pedestrians across different cameras lead to less satisfactory model performance. In contrast, ICS ReID benefits from more readily obtainable intra-camera annotated labels, which can significantly enhance the performance of ReID models.

\subsubsection{Comparison With Unsupervised Domain Adaptation Methods}
Finally, to further illustrate that person ReID with only intra-camera labels aligns better with real-world scenarios, we also compare our approach with the most representative state-of-the-art unsupervised domain adaptation (UDA) ReID methods. We report their best performance in Table \ref{tab:sotauda}. Among these methods, our approach surpasses the most advanced CaCL \cite{lee2023camera} by 22.4\% in terms of mAP on MSMT17 dataset. This significant improvement is primarily because UDA methods utilize supervised knowledge from the source domain to bridge the large domain gap in the target domain. In contrast, our approach relies solely on intra-camera labeled data in the target domain for learning, which is much less costly and achieves better results. These findings indicate that having more readily obtainable intra-camera annotation labels offers greater scalability.

\subsection{Ablation Studies}
In this section, we conduct detailed ablation experiments on each proposed component in the CCAFL. The results are shown in Table \ref{tab:xrsy}. 

\subsubsection{Baseline}
Compared to previous methods, ClusterContrast \cite{dai2022cluster} performs better by using contrastive learning and ID-centric memory. Therefore, in both Order 0 and Order 1, we establish corresponding ID-centric memory for intra- and inter-camera scenarios. Our baseline builds upon ClusterContrast \cite{dai2022cluster}, leveraging its efficient use of contrastive learning and feature centroid memory, resulting in notable accuracy improvements. It is worth mentioning that Order 1 is pre-trained on the CLIP-based model, and owing to its robust large-scale pre-training capability, it achieves better accuracy results compared to Order 0, which is pre-trained on ImageNet.

\subsubsection{Effectiveness of the Intra-camera Discriminative Learning}
Table \ref{tab:xrsy} shows that considering only the central features of IDs within the camera indeed yields considerable results. However, on the large-scale MSMT17 dataset, as the number of pedestrians within each camera increases, insufficient learning of these features can impact model performance. From Order 2, it can be seen that our designed ICDL method achieves an improvement of 1.3\% and 2.6\% in mAP on Market1501 and MSMT17, respectively, demonstrating its effectiveness. Therefore, properly considering global difficult negative samples can effectively encourage the model to learn fine-grained pedestrian features within each camera.

\subsubsection{Effectiveness of the Inter-camera Adversarial Learning}
The ICAL module is introduced to reduce differences in pedestrian feature distributions between cameras, addressing the shortcomings of cross-camera pseudo-labels. After directly incorporating the ICAL module in Order 5, the mAP performance on the MSMT17 dataset improves by 5.9\% compared to Order 1. Additionally, when combined with other modules, the ICAL module consistently enhances performance. This indicates that the ICAL module can effectively explore complex interactions across all sample pairs between cameras, further mitigating the interference caused by camera shifts.

\subsubsection{Effectiveness of the Image-Text Alignment Loss}
After comparing all learning components, we also verify the effectiveness of the textual information as a supervisory signal. Order 3 and Order 4 demonstrate the effectiveness of implicit text-based prompt learning, achieving a further 2.1\% improvement in mAP on the MSMT17 dataset compared to Order 2, indicating the effectiveness of the text description information provided by large-scale vision-language models.

\begin{figure}[t]
  \centering
  \begin{minipage}{0.5\textwidth}
    \centering
    \includegraphics[width=\textwidth]{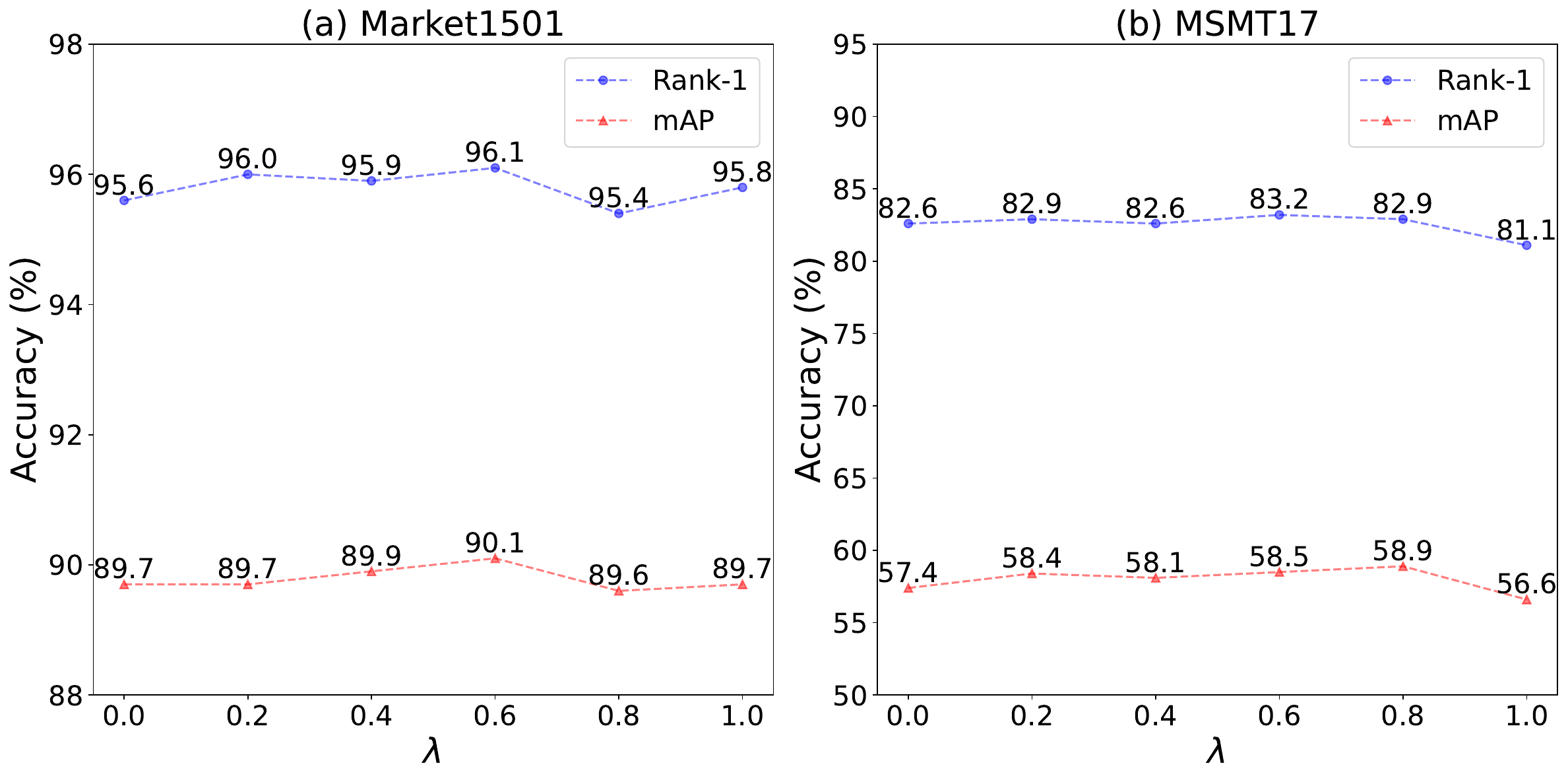}
    \caption{Impact of intra-camera loss weight $\lambda$ on Market-1501 and MSMT17 datasets.}
    \label{fig:fig1}
  \end{minipage}\hfill
  \begin{minipage}{0.5\textwidth}
    \centering
    \includegraphics[width=\textwidth]{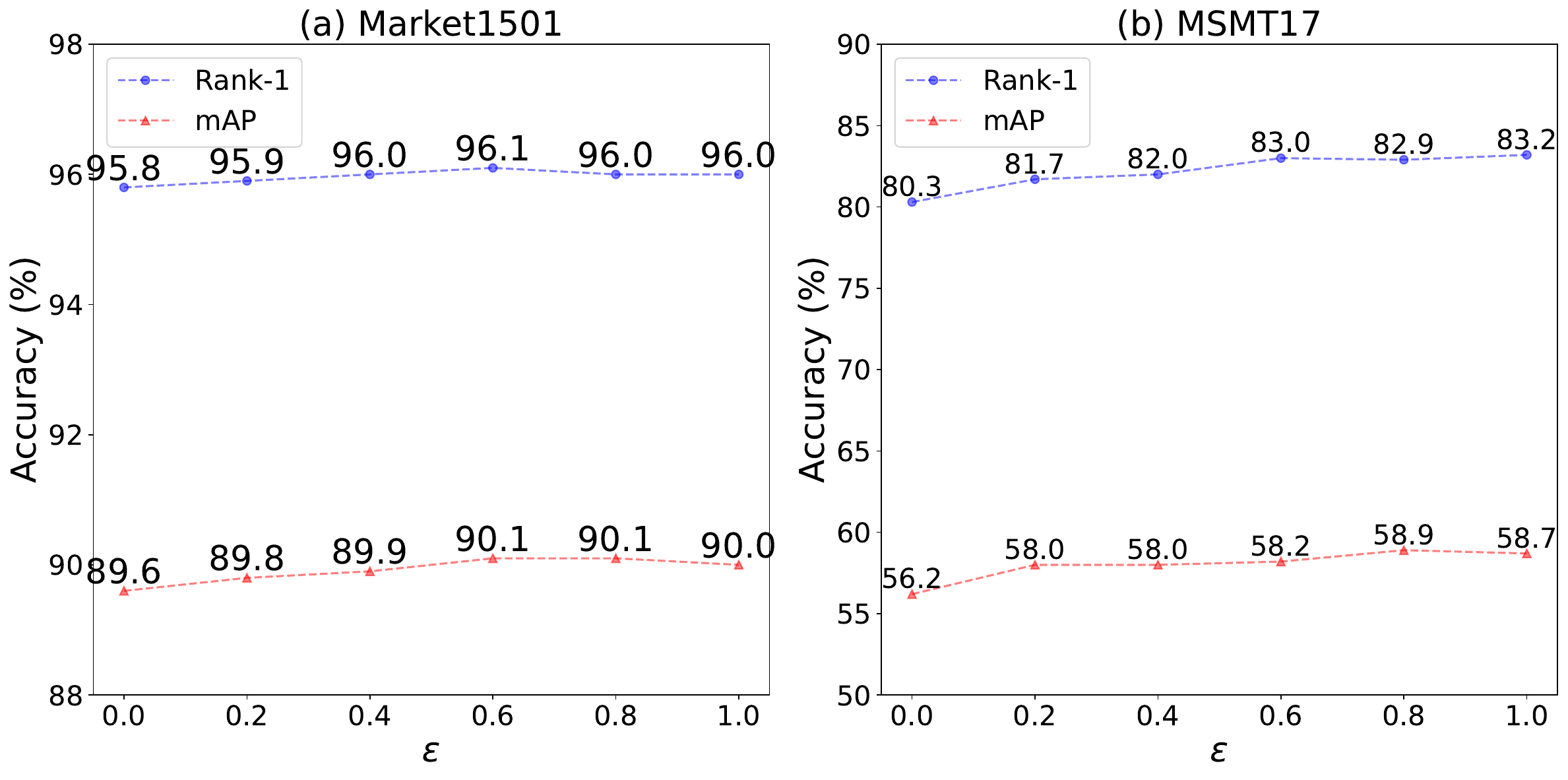}
    \caption{Impact of threshold $\epsilon$ on Market-1501 and MSMT17 datasets.}
    \label{fig:fig2}
  \end{minipage}
\end{figure}

\begin{figure}[t]
  \centering
  \begin{minipage}{0.5\textwidth}
    \centering
    \includegraphics[width=\textwidth]{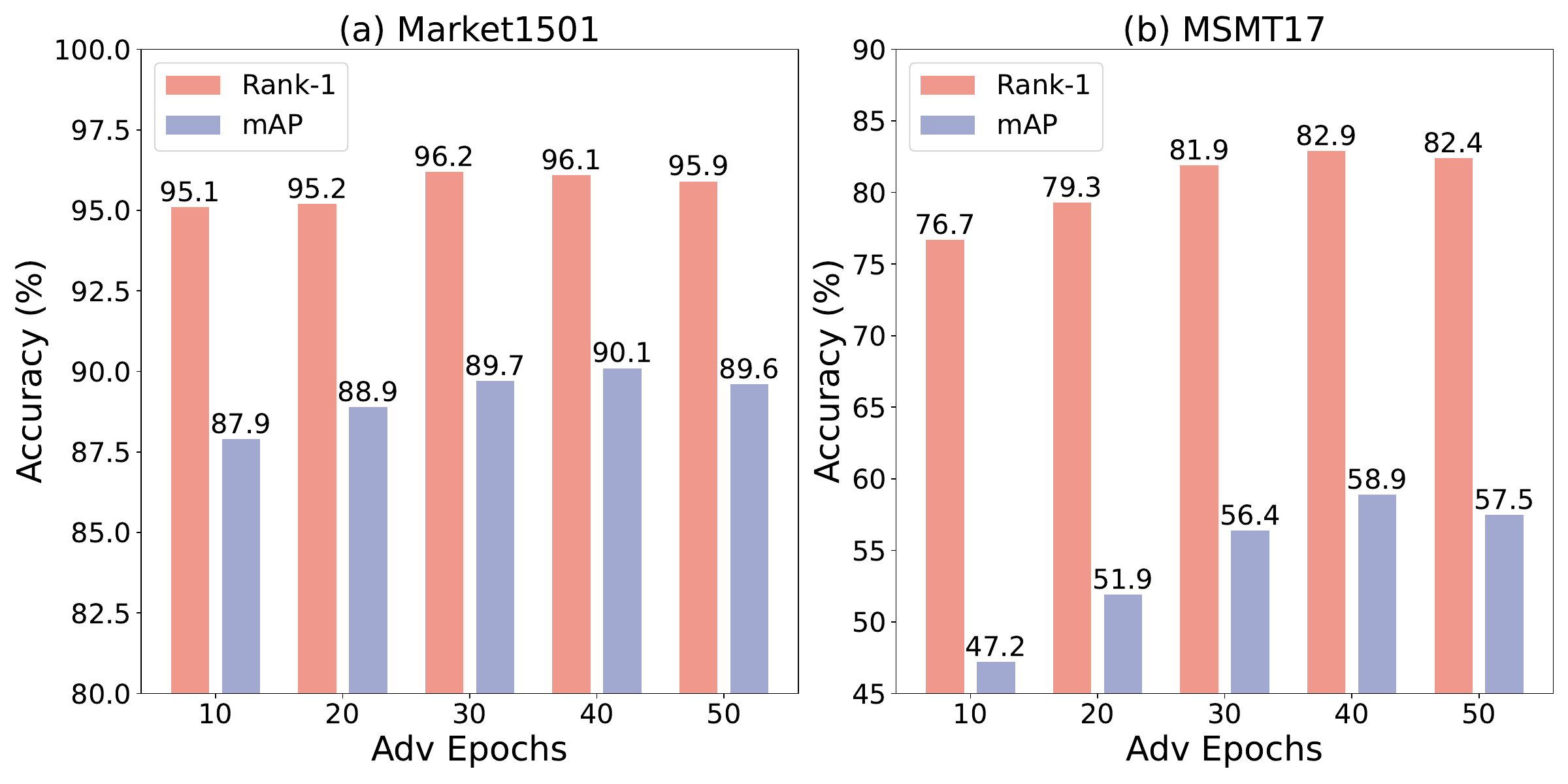}
\caption{Effect of starting epoch for inter-camera adversarial learning on Market-1501 and MSMT17 datasets.}
  \label{fig:fig3}
  \end{minipage}\hfill
  \begin{minipage}{0.5\textwidth}
    \centering
    \includegraphics[width=\textwidth]{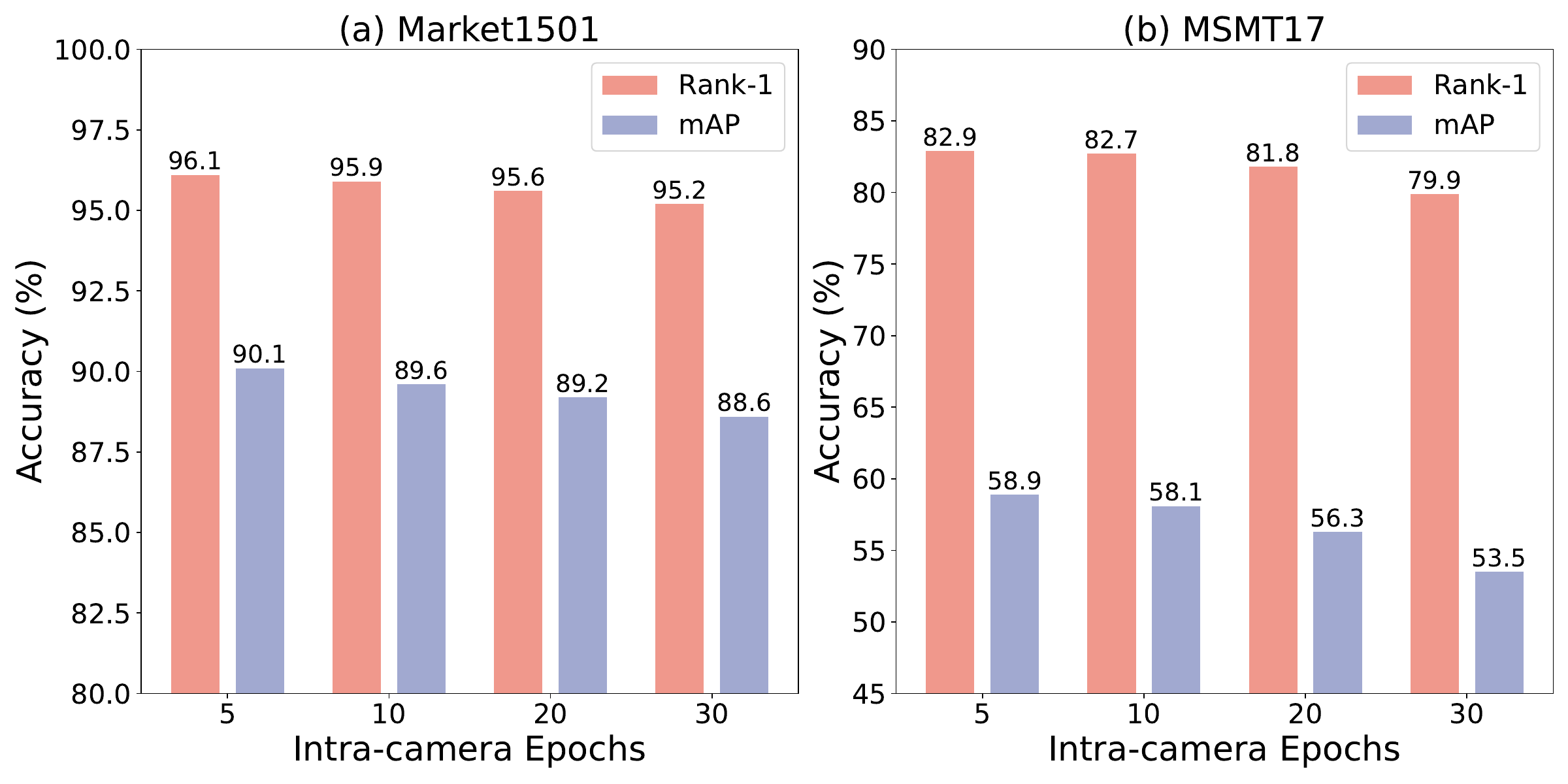}
  \caption{Effect of different intra-camera learning epochs on Market-1501 and MSMT17 datasets.}
  \label{fig:fig4}
  \end{minipage}
\end{figure}

\subsection{Parameter Analysis}
In this subsection, we analyze the sensitivity of the hyperparameters involved in CCAFL. We conduct experiments on the Market1501 and MSMT17 datasets to study the impact of these hyperparameters on model performance. Specifically, these hyperparameters include the balance weight $\lambda$ for intra-camera loss, the effectiveness of $\epsilon$ in inter-camera adversarial loss, the starting epoch for inter-camera adversarial learning, the distance threshold $T$ for inter-camera association, and the number of text tokens in the intra-camera prompting phase. Through these experiments, we aim to comprehensively evaluate the influence of each hyperparameter on the model's performance.

\subsubsection{The balance weight $\lambda$ for intra-camera loss}
We assess the impact of the balance weight parameter $\lambda \in [0,1]$ on the total loss of intra-camera learning, as shown in Eq. \ref{eq:07}. In Fig. \ref{fig:fig1}, we observe that the most appropriate $\lambda$ settings are 0.6 and 0.8 for the Market-1501 and MSMT17 datasets, respectively. This not only demonstrates that the proper utilization of intra-camera mean features and instance features leads to better performance improvements but also indicates that our proposed intra-camera discriminative loss helps the model learn finer-grained pedestrian features.

\subsubsection{The Effectiveness of $\epsilon$}
Fig. \ref{fig:fig2} shows the performance variation of CCAFL as $\epsilon$ changes from 0 to 1. This parameter primarily balances the model's inter-camera and intra-camera recognition abilities. When $\epsilon$ is larger, the model focuses more on inter-camera recognition. The figure illustrates that the optimal values for Market1501 and MSMT17 datasets are 0.8.

\subsubsection{The start epoch for inter-camera adversarial learning}
Table 5 shows the impact of the starting epoch for inter-camera adversarial learning on the model's performance. It is evident that when the classifier and adversarial training start simultaneously, the model performs worse than without ICAL. As the starting epoch for adversarial training increases, the classifier is better trained, and the global ID classifier becomes more capable of distinguishing images of the same person from different cameras. Consequently, the adversarial training becomes more effective. As shown in Fig. \ref{fig:fig3}, the optimal starting epoch for adversarial learning on Market1501 and MSMT17 datasets is 40, which validates the effectiveness of the adversarial approach.

 \begin{figure}[t]
  \centering
   \includegraphics[width=0.5\textwidth]{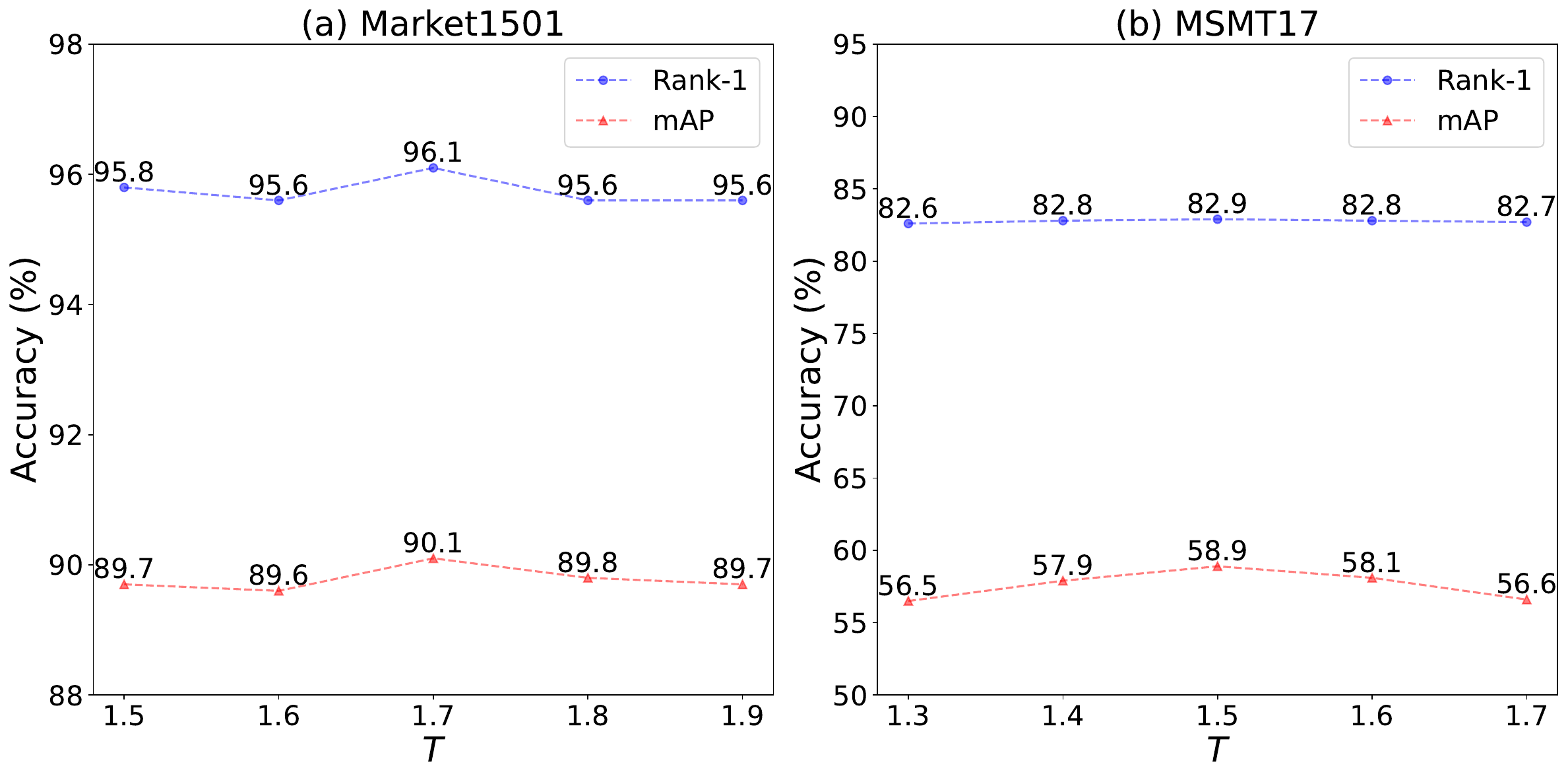}
  \caption{Impact of hyper-parameters threshold $T$ on Market-1501 and MSMT17 dataset.}
  \label{fig:associate}
\end{figure}

 \begin{figure}[t]
  \centering
   \includegraphics[width=0.5\textwidth]{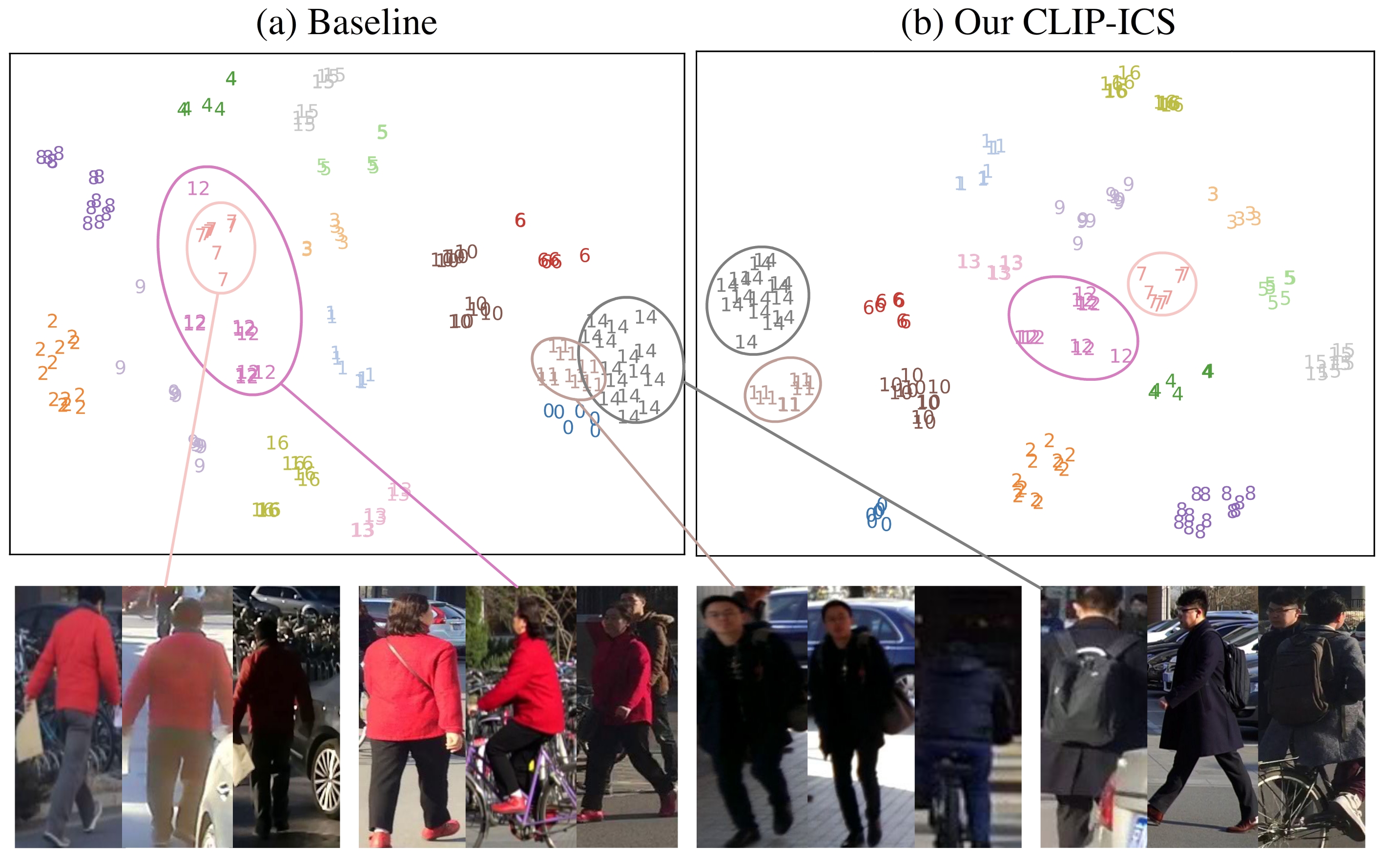}
  \caption{The T-SNE visualization presents the features acquired through different methods, arranged from left to right. The digits of identical colors indicate the same ID. We have circled and displayed below some representative example images with IDs \#7, \#11, \#12, and \#14.}
  \label{fig:tsne}
\end{figure}

\subsubsection{The Effectiveness of different intra-camera learning epochs}
Considering the poor quality of the representations learned by the model in the early training stage and the need to fully utilize the predefined label information within each camera, designing an intra-camera supervised learning phase is necessary to help the model warm up and learn pedestrian features under each camera. However, excessive intra-camera learning can lead to the model's inability to recognize inter-camera pedestrian features effectively. Therefore, we conduct experiments on the number of epochs for intra-camera learning. As shown in Fig. \ref{fig:fig4}, we observe that when the intra-camera learning epoch is set to 5, the performance is optimal across all datasets.

\subsubsection{The distance threshold $T$ for inter-camera association}
 We evaluate the distance threshold $T$ used for associating inter-camera IDs in Eq. \ref{eq:12}. The impact of different $T$ sizes on inter-camera-associated IDs is illustrated in Fig. \ref{fig:associate}. For Market-1501 and MSMT17, the model performance is optimized when $T$ is set to 1.7 and 1.5, respectively.

 \subsubsection{The Effectiveness of different Number of learnable tokens M}
Additionally, we analyze the number of learnable tokens \(M\) during the intra-camera predefined label prompting phase in Table \ref{tab:text-token}. The results indicate that the performance is optimal across all datasets when \(M = 5\).

 \subsubsection{Accuracy of inter-camera associated pseudo labels}
\begin{figure}[t]
  \centering
   \includegraphics[width=0.5\textwidth]{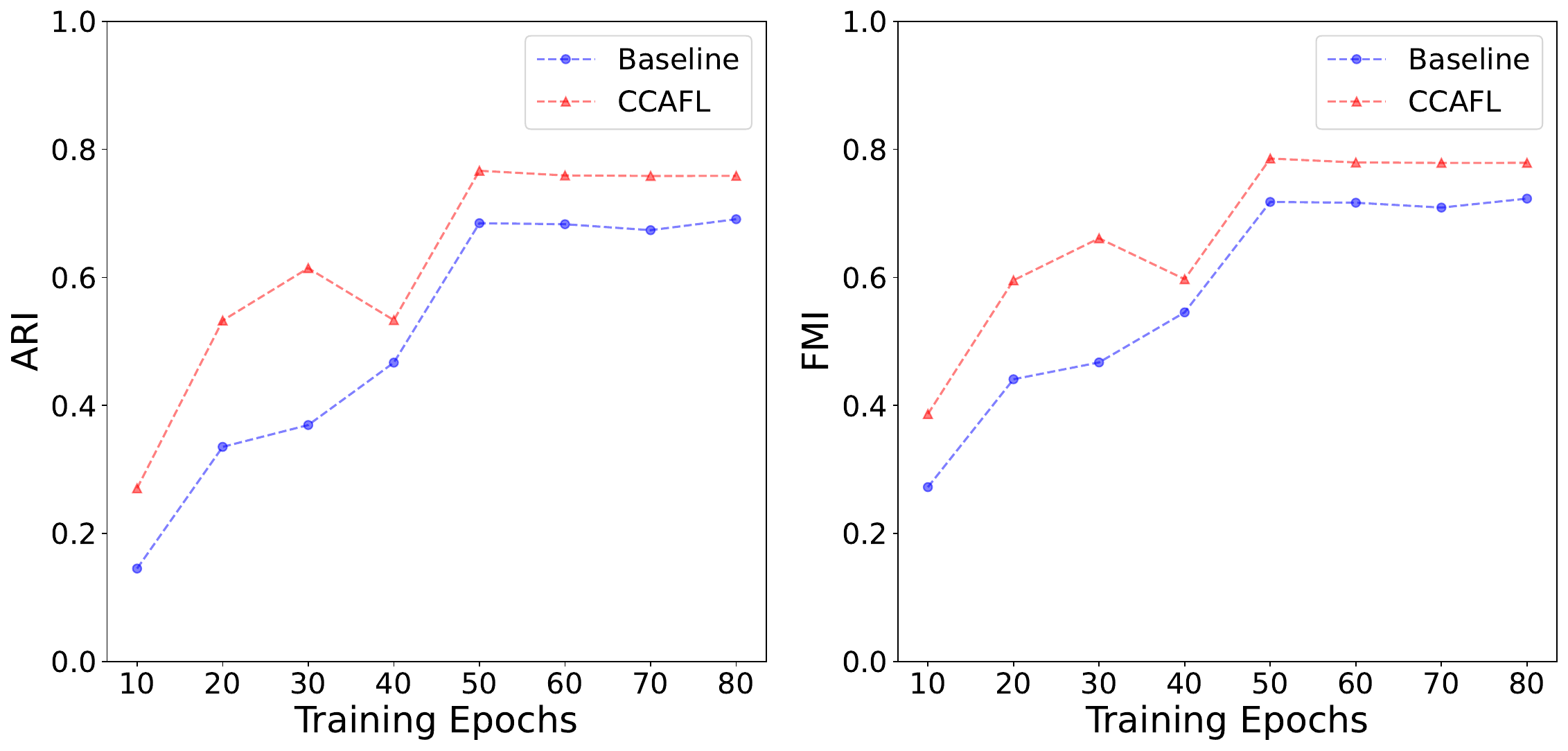}
  \caption{Comparison of inter-camera clustering quality across different epochs between Baseline and CCAFL on Market1501.}
  \label{fig:ari}
\end{figure}

Fig. \ref{fig:ari} further demonstrates the effectiveness of our method. We compare our proposed approach with the baseline model using two common clustering evaluation metrics. The results show that the camera-invariant pseudo-labels generated by CCAFL significantly outperform the baseline model on both metrics. This indicates that by learning camera-invariant features, our method provides more reliable pseudo-labels for network training, thus significantly improving performance

\begin{table}[h]
\caption{Analysis of Different Numbers of Learnable Tokens (M)}
\centering
\label{tab:text-token}
\resizebox{0.30\textwidth}{!}{ % 将宽度调整为页面宽度的一半
\begin{tabular}{c|cc|cc}
\hline
\multirow{2}{*}{{M}} & \multicolumn{2}{c|}{Market-1501}   & \multicolumn{2}{c}{MSMT17}    \\ \cline{2-5} 
                   & mAP           & R1            & mAP           & R1            \\ \hline
2                  & 89.6          & 95.4          & 57.8          & 82.4          \\ \hline
4                  & 89.7          & 95.6          &  55.5             &  81.7             \\ \hline
5                  & \textbf{90.1} & \textbf{96.1} & \textbf{58.9} & \textbf{82.9} \\ \hline
6                  & 89.9          & 95.9          &  58.2              &  82.7             \\ \hline
\end{tabular}}
\end{table}

\subsection{Visualization}
We separately perform feature visualization on the baseline (Order 1) and Order 7 on MSMT17 dataset using T-SNE \cite{van2014accelerating}, as shown in Fig. \ref{fig:tsne}.
% In Fig. \ref{fig:tsne} (a), which depicts the model with only the pretrained weights loaded, the model struggles to accurately recognize each pedestrian, resulting in a highly scattered distribution. 
Fig. \ref{fig:tsne} (a), representing the baseline model, demonstrates improved ability in distinguishing different pedestrian identities. However, its recognition performance deteriorates when confronted with IDs such as ID\#11 and ID\#14, which have substantial background noise, and the highly similar ID\#7 and ID\#12. In contrast, as shown in Fig. \ref{fig:tsne} (b), our model effectively differentiates these challenging IDs and ensures a higher concentration of pedestrian features across multiple cameras, thereby confirming the inherent superiority of our approach in cross-camera pedestrian identification.

% In which, image (a), serving as the baseline, exhibits competence in identifying distinct pedestrian identities accurately. However, its recognition efficacy shows signs of exacerbation when confronted with ID\#11 and ID\#14, laden with substantial background noise, and the markedly similar ID\#7 and ID\#12. Contrarily, as illustrated in image (b), our model effectively distinguishes these intricate IDs and ensures a higher concentration of pedestrian features captured across varied cameras, thereby affirming our approach's inherent superiority in cross-camera pedestrian identification.

\section{Conclusion}
This paper explores the application of CLIP in the semi-supervised task of ICS ReID and proposes a three-stage training strategy for CLIP-based camera-invariant feature learning. First, we utilize the annotated labels within each camera to extract the implicit semantic information of pedestrians, providing supervisory information for subsequent intra-camera and inter-camera learning. Second, during the intra-camera learning stage, we design a discriminative contrastive loss based on hard negative mining, effectively increasing the variance among different pedestrians within the same camera, which helps the model learn finer-grained pedestrian features. Then, in the inter-camera learning stage, we combine the semantically rich information extracted in the first stage with inter-camera prototype contrastive learning to learn pedestrian features that vary across viewpoints. Finally, to further reduce sensitivity to noise and enhance the model’s adaptability to complex real-world scenarios, we design an inter-camera adversarial learning method to minimize the discrepancies in pedestrian features between cameras, thereby reducing the impact of camera bias. Extensive experiments demonstrate that our method outperforms state-of-the-art methods on three commonly used person ReID datasets, showcasing its superiority.

\bibliographystyle{IEEEtran} % 设置参考文献样式
\bibliography{Reference.bib} % 指定参考文献文件


\end{document}